\documentclass[letterpaper]{article} 
\usepackage{aaai23}  
\usepackage{times}  
\usepackage{helvet}  
\usepackage{courier}  
\usepackage[hyphens]{url}  
\usepackage{graphicx} 
\usepackage{natbib}  
\usepackage{caption} 
\usepackage{amsfonts}
\usepackage{amsmath}
\usepackage{bm}
\usepackage{bbold}
\usepackage{cuted}
\usepackage{lipsum}
\usepackage{mathtools}

\title{Maximum Likelihood on the Joint (Data, Condition) Distribution for Solving Ill-Posed Problems with Conditional Flow Models}
\author {
    John S. Hyatt
}
\affiliations{
    Computational and Information Sciences Directorate\\
    DEVCOM Army Research Laboratory\\
    Aberdeen Proving Ground, Maryland 21005\\
    john.s.hyatt11.civ@army.mil
}

\nocopyright

\begin{document}

\maketitle

\begin{abstract}
In the absence of conditioning information, normalizing flow models are trained using maximum likelihood. Previous extensions to conditional settings have abandoned the simple maximum-likelihood approach, as obtaining the full joint distribution of data and conditions for training purposes (or equivalently, the true conditional distribution for each point in the training dataset) is impossible. However, under special conditions that nevertheless apply to a wide class of useful problems, it is possible to obtain a simple, computationally cheap proxy with the same global minimum as the conditional likelihood. Conditional flow models trained using this proxy do not require explicit knowledge of the conditional distribution, auxiliary networks or other specific architecture, or additional loss terms beyond maximum likelihood. They provide a natural and powerful way to solve a class of ill-posed inverse problems. I demonstrate these properties on easily visualized toy problems, then use the method to successfully generate class-conditional images and to reconstruct semantically degraded images via super-resolution.
\end{abstract}

\section{Introduction}
\label{sec:introduction}

Consider the problem of reconstructing an image degraded by blurring, low resolution, occlusion, or other effects. If the scale of degradation is small compared to the features in the image, interpolation with some basic prior knowledge about image properties is sufficient to recover the original with a high degree of certainty. On the other hand, if the image is so damaged that small- and medium-scale features are totally lost, the problem is \textit{ill-posed} and, instead of a single solution, has a highly complex \textit{distribution} of possible reconstructions consistent with the degraded input, each with a different set of synthesized replacement features.

For aesthetic applications, it may be sufficient only to require one realistic reconstruction. However, applications like uncertainty quantification or risk assessment require the full conditional distribution of possible reconstructions along with a measure of trustworthiness (how well the predicted distribution matches the unknown true distribution).

Flow models are capable of learning and sampling from arbitrary data distributions. Previous attempts to model conditional probability distributions with flow models have sacrificed some of the properties that make them attractive in the first place, such as interpretability, bijectivity, or a simple loss function. I am aware of one method for conditioning flow models that retains all of these properties; in this paper I develop a new approach, complementary to that one, and much simpler for a broad class of problems.

Formally, if $\mathcal{X}$ is the space of all possible images in the target domain and $\mathcal{Y}$ the space of degraded images in the same domain, each point $\mathbf{y}\in\mathcal{Y}$ defines a conditional distribution, $p_{X\vert Y=\mathbf{y}}(\mathbf{x})$, of points $\mathbf{x}\in\mathcal{X}$. A non-conditional flow model learns to sample from the distribution $p_X(\mathbf{x})$; this paper describes a way to train a flow model on the joint distribution $p_{XY}(\mathbf{x},\mathbf{y}) = p_{X\vert Y=\mathbf{y}}(\mathbf{x})p_Y(\mathbf{y})$ using a simple, accurate, computationally cheap proxy for $p_Y(\mathbf{y})$. Specifying $\mathbf{y}$ then gives the conditional distribution $p_{X\vert Y=\mathbf{y}}(\mathbf{x})$. The resulting conditional model is a direct extension of non-conditional flow models, so it retains all their desirable properties: efficient sampling and inference, a well-behaved and meaningful latent space, and exact reconstruction of encoded latent vectors. It also retains the same weaknesses, such as low model expressivity and difficulty in accurately assigning likelihood to new data, which remain open problems in flow model design.

\section{Related Work}
\label{sec:related_work}

Generative modeling has been the subject of intensive research for much of the last decade, resulting in a deep and diverse catalog of approaches. The need to guarantee good sampling from an arbitrary data distribution has driven a movement from conventional models towards density models such as normalizing flows.

\subsection{Generative Modeling with GANs and VAEs}
\label{subsec:non_prob_generative_model}

Historically, the most popular generative machine learning models have been based on Generative Adversarial Networks (GANs, \citeauthor{GANs} \citeyear{GANs}) and Variational Autoencoders (VAEs, \citeauthor{VAEs} \citeyear{VAEs})\footnote{This paper was originally written before the advent of VLMs and modern generative AI. However, many of the stated problems persist; in general, these models rely on brute-force universal function approximation and provide no guarantees.}. Both learn a map $G : Z \rightarrow X$ from a low-dimensional latent space $\mathcal{Z}$ to a high-dimensional data space $\mathcal{X}$. Since their introductions, GANs and VAEs have benefited from an enormous amount of research (for surveys, see \citeauthor{GAN_survey_1} \citeyear{GAN_survey_1} and \citeauthor{GAN_survey_2} \citeyear{GAN_survey_2}). However, GANs and VAEs generally have some undesirable properties.

\subsubsection{Non-likelihood Basis}
\label{subsubsec:non_likelihood}

Ideally, training should \textit{maximize the likelihood} that $G$ samples from the true data distribution $p_X(\mathbf{x})$; or, for a conditional model, from $p_{X\vert Y=\mathbf{y}}(\mathbf{x})$, given some $\mathbf{y}$. VAEs maximize a lower bound on the likelihood, while the adversarial loss used to train GANs does not incorporate likelihood at all. Flow models maximize likelihood directly, as described in Section\ \ref{subsec:prob_generative_model}.

\subsubsection{$G(\mathcal{Z})$ and $\mathcal{X}$ Have Disjoint Support}
\label{subsubsec:disjoint_support_Z_X}

The space of generated examples $G(\mathcal{Z})$ and the data space $\mathcal{X}$ are both low-dimensional manifolds embedded in a much higher-dimensional space \citep{low_dim_manifold_data,noise_injection_GANs}. These manifolds have disjoint support, which causes $G$ to receive bad gradients during training. Solutions to this problem include revising the objective function \citep{OriginalWGAN,ImprovedWGAN,InfoGAN,infoVAE}, or adding noise to both $\mathcal{X}$ and $G(\mathcal{Z})$ \citep{instance_noise}.

In contrast, flow models are bijections between $\mathcal{Z}$ and $\mathcal{X}$. Simple noise annealing and data augmentation are enough to ensure overlapping support, as discussed in Section \ref{subsubsec:overlapping_support}.

\subsubsection{$G$ is Not Invertible}
\label{subsubsec:non_inv_generative_model}

For GANs and VAEs, $\dim(\mathcal{Z})<\dim(\mathcal{X})$, so $G^{-1} : X \rightarrow Z$ is not well-defined. If $G$ is a deep neural network, it is also highly nonlinear and therefore not invertible in practice even if $\dim(\mathcal{Z})=\dim(\mathcal{X})$.

Interpolating from $\mathbf{z}_1$ to $\mathbf{z}_2$ in $\mathcal{Z}$ provides a way to interpolate from $G(\mathbf{z}_1)$ to $G(\mathbf{z}_2)$ in $G(\mathcal{Z})$; if $\mathbf{z}_1$ and $\mathbf{z}_2$ correspond to a particular feature axis in the generated data, this can be used to manipulate that data \citep{z_feature_arithmetic}. In order to manipulate a real data example $\mathbf{x}$ in this way, it must be first mapped to $\mathcal{Z}$; if $G$ is not invertible, there is no way to obtain a latent vector corresponding to $\mathbf{x}$. Inverting $G$ typically means either learning an encoder $E\approx G^{-1}$ or iteratively updating $\mathbf{z}$ to obtain $G(\mathbf{z})=\mathbf{x}$ for some target $\mathbf{x}$ (for a survey, see \citeauthor{GAN_inversion} \citeyear{GAN_inversion}).

In contrast, flow models are trivially invertible, providing direct access to latent space. The model is a bijective map between $\mathcal{Z}$ and $\mathcal{X}$: then $\mathbf{z}$ and $\mathbf{x}$ are equally good representations because they contain the same information.

\subsubsection{Conditional GANs and VAEs}
\label{subsubsec:cond_models_other_mods}

These combine prior information and random sampling to learn a map $G : (Z,Y) \rightarrow X$. Those capable of \textit{realistic} and \textit{diverse} predictions can be very complex (see, for example, \citeauthor{BicycleGAN}, \citeyear{BicycleGAN}), but their predictions are not guaranteed to be \textit{representative} of the true data distribution, making them ill-suited for quantitative analysis. Meanwhile, inverting $G$ is complicated by the fact that $G^{-1}(\mathbf{x})$ must disentangle $\mathbf{y}$ and $\mathbf{z}$.

The conditional flow model described in this paper is conceptually simple, learning a map $F : (X,Y) \rightarrow (Z,Y)$. Because it maximizes likelihood, its predictions represent the true target distribution to within the capabilities of the chosen model architecture; because it is trivially invertible, the quality of the learned map can be evaluated at low cost using held-out validation data, as shown in Section\ \ref{subsec:histograms}.

\subsection{Likelihood-Based Generative Modeling with Non-conditional Normalizing Flows}
\label{subsec:prob_generative_model}

Normalizing flow models sidestep most of the problems described in Section\ \ref{subsec:non_prob_generative_model}. A non-conditional flow trains on maximum likelihood using change of variables to learn a bijective map $F : X \rightarrow Z$ between data space $\mathcal{X}$ and latent space $\mathcal{Z}$ \citep{NICE,RealNVP}. The latent prior, $p_Z$, is related to the probability density function of the data, $p_X$, by
\begin{equation}
\label{eq:change_of_variables_XZ}
    p_X(\mathbf{x})
    =
    p_Z(F(\mathbf{x}))
    \left\vert
        \det
        \mathbb{J}_F(\mathbf{x})
    \right\vert,
\end{equation}
where $\mathbb{J}_F(\mathbf{x}) = \partial F(\mathbf{x}) / \partial \mathbf{x}^\mathsf{T}$ is the Jacobian of $F$. The training objective of a non-conditional flow model $F$ is
\begin{equation}
\begin{aligned}
\label{eq:log_change_of_variables_XZ}
    \mathcal{L}&^\textrm{Flow}(F)
    =
    \mathbb{E}_{\mathbf{x}\sim p_X^\textrm{data}(\mathbf{x})}
    \left[
    -
    \log
    p_X(\mathbf{x})
    \right]
    \\
    &=
    \mathbb{E}_{\mathbf{x}\sim p_X^\textrm{data}(\mathbf{x})}
    \left[
    -
    \log 
    p_Z(F(\mathbf{x}))
    -
    \log
    \left\vert 
        \det \mathbb{J}_F(\mathbf{x})
    \right\vert
    \right].
\end{aligned}
\end{equation}
Once the model is trained, it can then be used to sample $\mathbf{x} \sim p_X(\mathbf{x})$ by evaluating $F^{-1}(\mathbf{z})$ with $\mathbf{z}\sim p_Z(\mathbf{z})$, or to map new data into latent space as $F(\mathbf{x})$.

The most popular flow model is Real NVP \citep{RealNVP}, which composes multiple simple neural-network-based affine scaling transformations. More powerful models include Glow\ \citep{INNs_glow}, Flow++\ \citep{INNs_flowplusplus}, invertible convolutional flow\ \citep{inv_conv_flow}, and autoregressive flows \citep{IAFs,MAFs}. All the experiments in this paper are based on Real NVP, but the conditional described is agnostic to the details of $F$.

It is important to note that flow models do not solve every problem with generative modeling, as discussed briefly below. These are open research problems.

\subsubsection{Infinitely Many Valid Maps}
\label{subsubsec:infinite_maps}

Eq.\ \ref{eq:log_change_of_variables_XZ} does not have a unique global optimum \citep{flow_for_OOD_1}. Consider the case where both the prior and target distributions are a two-dimensional standard normal Gaussian: $p_Z(\mathbf{z}) = p_X(\mathbf{x}) = \mathcal{N}(\mathbf{0},\mathbf{1})$. The simplest answer is that $F$ is the identity map, but from the point of view of Eq.\ \ref{eq:log_change_of_variables_XZ}, infinitely many $F$ (rotations, reflections, etc.) are equally optimal.

In terms of local minima, the same likelihood can also be achieved for many $F$, as long as the sum of both terms in Eq.\ \ref{eq:log_change_of_variables_XZ} remains constant. A well-optimized $F$ will output the correct distribution, but without the right inductive biases, the learned map probably operates on undesirable principles \citep{density_models_anomaly_detection}.

\subsubsection{Inappropriate Inductive Biases}
\label{subsubsec:inductive_biases}

Normalizing flows have been shown to learn local pixel correlations rather than capture the semantic features of their training data \citep{flow_for_OOD_2}. As a result, of the many $F$ with similar $\mathcal{L}^\textrm{Flow}$, the model will learn one that assigns low probability to inputs with unusual local correlations, rather than unusual semantic content. $F$ may therefore not be able to correctly detect inputs with anomalous semantics if it is simply trained according to Eq.\ \ref{eq:log_change_of_variables_XZ}, which can also lead to sensitivity to adversarial attacks \citep{flow_adversarial}. Ensuring $F$ has the correct inductive bias is an open problem in flow model design, and is compounded for conditional flow models, as I show in Section\ \ref{subsec:continuous_toy_problem}.

\subsection{Modeling Conditional Distributions with Parameterized Flow}
\label{subsec:cond_flow_models}

In a bijective map, the input and output must have the same dimensionality and contain the same information. In a non-conditional flow, these are $\mathbf{x}$ and $\mathbf{z}$. For a conditional problem, the goal is to sample $\mathbf{x} \sim p_{X\vert Y=\mathbf{y}}(\mathbf{x})$ given $\mathbf{y} \in \mathcal{Y}$, so $\mathbf{x}$ is determined by information in both $\mathbf{y}$ and $\mathbf{z}$. What is the correct relationship between $\mathbf{x}$, $\mathbf{y}$, and $\mathbf{z}$?

Many conditional flows parameterize a map on the condition as $F_Y : X \rightarrow Z$, which can be as simple as learning a different non-conditional model for each of a set of discrete conditions $y\in(y_1,y_2,\dots,y_N)$, and choosing which to apply based on the condition. The alternative is to parameterize the prior as $p_{Z\vert Y=\mathbf{y}}$. In both cases, this parameterization is typically at least partially non-invertible.

Prenger et al. (\citeyear{cINNs_10}), Lu and Huang (\citeyear{cINNs_6}), Rombach et al. (\citeyear{cINNs_4}), Ardizzone et al. (\citeyear{cINNs_2}), and Anantha Padmanabha and Zabaras (\citeyear{cINNs_3}) parameterize $F$ by embedding $\mathbf{y}$ into its internal layers, either directly or via an auxiliary network; additional terms must be added to the objective to ensure $F$ has the correct parameterization. The parameterized map is invertible with respect to $Z$ and $X$, but the actual process of parameterizing $F$ is still non-invertible. Notably, because $\mathbf{y}$ is not part of the flow, it must be embedded separately into \textit{every transformation} in $F$. Winkler et al. (\citeyear{cINNs_5}) follow this approach and additionally parameterize the prior.

Liu et al. (\citeyear{cINNs_1}) also parameterize the prior with an encoder network $E$ that non-invertibly combines $\mathbf{y}$ and $\mathbf{z}$ into one latent representation $\mathbf{z}^*=E(\mathbf{z},\mathbf{y})$, with additional loss terms to force $\mathbf{z}^*$ to be similar to $F(\mathbf{x})$.  $F$ is an invertible map, so no repeated embedding is necessary; however, given a new example $\mathbf{x}$, the model can only obtain $\mathbf{z}^*$, with no way to disentangle $\mathbf{z}$ from $\mathbf{y}$. As a result, this method can only be used to generate data during inference, not map new data back into the original latent space.

Ardizzone et al. (\citeyear{cINNs_8}) use a Gaussian mixture model as their prior, maximizing the mutual information between the class labels and Gaussian clusters in latent space. The model then classifies inputs based on the region of latent space containing $F(\mathbf{x})$. Sorrenson et al. (\citeyear{cINNs_9}) perform a similar parameterization based on independent component analysis.

To my knowledge, the only conditional flow method that is fully invertible and trained using only maximum likelihood is the recent work of \citeauthor{cINNs_7} (\citeyear{cINNs_7}); they also optimize on the joint distribution, albeit with a different approach. In an extension of the original flow design, they couple two bijective maps: a non-conditional model $F' : Y \rightarrow Z'$ that learns $p_Y$, and a conditional model $F : X \rightarrow (Z,Z')$ that learns $p_{X\vert Y=\mathbf{y}}$. Together, they model the joint distribution $p_{XY}(\mathbf{x},\mathbf{y})=p_{X\vert Y=\mathbf{y}}(\mathbf{x})p_Y(\mathbf{y})$. Layers in $F$ operate on both the output of the previous layer in $F$ and that of the corresponding layer in $F'$. During inference, $F'$ encodes $\mathbf{y}$ into its own latent representation, while $\mathbf{z}$ is sampled from the latent prior. The two latent vectors are then concatenated, and $\mathbf{x}\sim p_{X\vert Y=\mathbf{y}}(\mathbf{x})$ is generated as $F^{-1}(\mathbf{z},F'(\mathbf{y}))$. Their approach, where $p_Y$ is directly modeled by $F'$, is orthogonal to the one described in this paper, where it is \textit{implicit} in the loss function.

\section{Modeling the Joint Distribution With Implicitly Parameterized Flow}
\label{sec:joint_flow_maps}
A fully symmetric conditional map must have conditions associated to both the input and output, such that the total information on both sides of the map is constant. Suppose $Y$ is the condition associated with $X$, and $Y'$ the condition associated with $Z$. If $X$ and $Y$ contain the same information as $Z$ and $Y'$, the bijective map $F : (X,Y) \rightarrow (Z,Y')$ preserves information symmetry and invertibility:
\begin{subequations}
\begin{align}
\label{eq:cINN_model}
    (\mathbf{x},\mathbf{y})
    =
    F^{-1}(\mathbf{z},\mathbf{y}')
    &=
    (F_X^{-1}(\mathbf{z},
        \mathbf{y}'),
     F_Y^{-1}(\mathbf{z},
        \mathbf{y}'))
    \\
\label{eq:cINN_inverse_model}
    (\mathbf{z},\mathbf{y}')
    =
    F(\mathbf{x},\mathbf{y})
    &=
    (F_Z(\mathbf{x},
        \mathbf{y}),
     F_{Y'}(\mathbf{x},
        \mathbf{y}))
    .
\end{align}
\end{subequations}
$F_X^{-1}$ and $F_Y^{-1}$ are the $X$ and $Y$ components of $F^{-1}$, and $F_Z$ and $F_{Y'}$ the $Z$ and $Y'$ components of $F$, respectively. $F$ is otherwise similar to the non-conditional case described in Section\ \ref{subsec:prob_generative_model}. This formulation has desirable properties: $F$ is fully invertible with respect to both its inputs; the components $(X,Y)$ and $(Z,Y')$ are always separable; and sampling and density estimation each require only one pass.

In general, the relationship between $Y$ and $Y'$ is ambiguous, although the total information in $(X,Y)$ and $(Z,Y')$ must be equal. In the special case where $Y = Y'$, $F$ defines a fully invertible conditional flow map that can be trained on maximum likelihood and has no need for additional loss terms or non-invertible model components.

\subsection{Change of Variables for a Joint Distribution}
\label{subsec:change_variables_joint}
Analogous to Eq.\ \ref{eq:change_of_variables_XZ}, the change of variables for this map is
\begin{align}
\label{eq:change_of_variables_XYZY}
    p_{XY}(\mathbf{x},\mathbf{y})
    &=
    p_{ZY'}(F(\mathbf{x},\mathbf{y}))
    \lvert
    \det \mathbb{J}_F(\mathbf{x},\mathbf{y})
    \rvert,
\end{align}
where $p_{XY}(\mathbf{x},\mathbf{y})=p_{X\vert Y=\mathbf{y}}(\mathbf{x})p_Y(\mathbf{y})$ and $p_{ZY'}(\mathbf{z},\mathbf{y}')=p_Z(\mathbf{z})p_{Y'}(\mathbf{y}')$ are joint probability densities and $\mathbb{J}_F(\mathbf{x},\mathbf{y}) = \partial F(\mathbf{x},\mathbf{y}) / \partial [\mathbf{x},\mathbf{y}]^\mathsf{T}$ is the Jacobian. The latent variable $Z$ is sampled independently from $Y'$; as shown in Section\ \ref{sec:experimental_results}, $F$ encodes universal features into $\mathcal{Z}$.

Taking the negative log, Eq.\ \ref{eq:change_of_variables_XYZY} can be rewritten as
\begin{equation}
\label{eq:log_change_of_variables_XYZY}
\begin{aligned}
    -\log p_{XY} (\mathbf{x},\mathbf{y})
    =&
    -\log p_{Z}(F_Z(\mathbf{x},\mathbf{y}))
    \\
    &
    -\log p_{Y'}(F_{Y'}(\mathbf{x},\mathbf{y}))
    \\
    &
    -\log \lvert
    \det \mathbb{J}_F(\mathbf{x},\mathbf{y})
    \rvert.
\end{aligned}
\end{equation}
The first and third terms are identical to the two terms in Eq.\ \ref{eq:log_change_of_variables_XZ}, except that they now also depend on $\mathbf{y}$. The second term, containing $p_{Y'}$, does not have a non-conditional counterpart.

$F$ cannot be optimized on Eq.\ \ref{eq:log_change_of_variables_XYZY} directly without knowing $p_{Y'}$. If $p_{Y'}$ is assumed to have a certain form, this term can be written explicitly: for example, if $p_{Y'}$ is Gaussian with covariance matrix $\mathbb{\Sigma}\propto\mathbb{1}$, then $-\log p_{Y'}(F_{Y'}(\mathbf{x},\mathbf{y}))=\frac{1}{2\sigma}\Vert F_{Y'}(\mathbf{x},\mathbf{y})-\mathbf{y}')\Vert_2$, up to a constant term. $p_{Y'}$ can also be modeled with a non-conditional flow, as in \citeauthor{cINNs_7}

Alternatively, it is sufficient to note that, regardless of the form of $p_{Y'}$, the only effect of the second term in Eq.\ \ref{eq:log_change_of_variables_XYZY} is to minimize the difference between $F_{Y'}(\mathbf{x},\mathbf{y})$ and $\mathbf{y}'$. However, the same is true of a simple proxy function such as any measure $d(F_{Y'}(\mathbf{x},\mathbf{y}),\mathbf{y}')$ of the distance between $F_{Y'}(\mathbf{x},\mathbf{y})$ and $\mathbf{y}'$. The optimization path will be different, but the final result will be the same as if the model had optimized on Eq.\ \ref{eq:log_change_of_variables_XYZY}, although the true $p_Y$ and $p_{Y'}$ are both still unknown.

Finally, for a broad class of problems, $Y=Y'$, allowing $p_Y$ and $p_{Y'}$ cancel in Eq.\ \ref{eq:change_of_variables_XYZY},  to give a formula for $p_{X\vert Y=\mathbf{y}}(\mathbf{x})$. Conveniently, this can be incorporated into $d$ to give the conditional training objective:
\begin{equation}
\label{eq:final_training_objective}
\begin{aligned}
    \mathcal{L}^\textrm{cFlow}
    =
    \mathbb{E}_{(\mathbf{x},\mathbf{y})\sim p_{XY}^\textrm{data}(\mathbf{x},\mathbf{y})}
    \big[
    &-\log p_{XY}(\mathbf{x},\mathbf{y})
    \big]
    \\
    =
    \mathbb{E}_{(\mathbf{x},\mathbf{y})\sim p_{XY}^\textrm{data}(\mathbf{x},\mathbf{y})}
    \big[
    &-\log p_Z(F_Z(\mathbf{x},\mathbf{y}))
    \\
    &+
    \lambda\cdot d(F_{Y'}(\mathbf{x},\mathbf{y}),\mathbf{y})
    \\
    &-\log \lvert \det \mathbb{J}_F(\mathbf{x},\mathbf{y})\rvert
    \big],
\end{aligned}
\end{equation}
where $\lambda$ just has to be large enough that the second term dominates the loss at the start of training.\footnote{A similar term is manually added to Eq. 14 of \citeauthor{cINNs_7} (\citeyear{cINNs_7}) to reinforce the importance of low-frequency cycle consistency between ground truth $\mathbf{x}$ and generated $F^{-1}_X(\mathbf{z},\mathbf{y}')$.}

\begin{figure}[t]
\centering
\includegraphics[width=0.75\columnwidth]{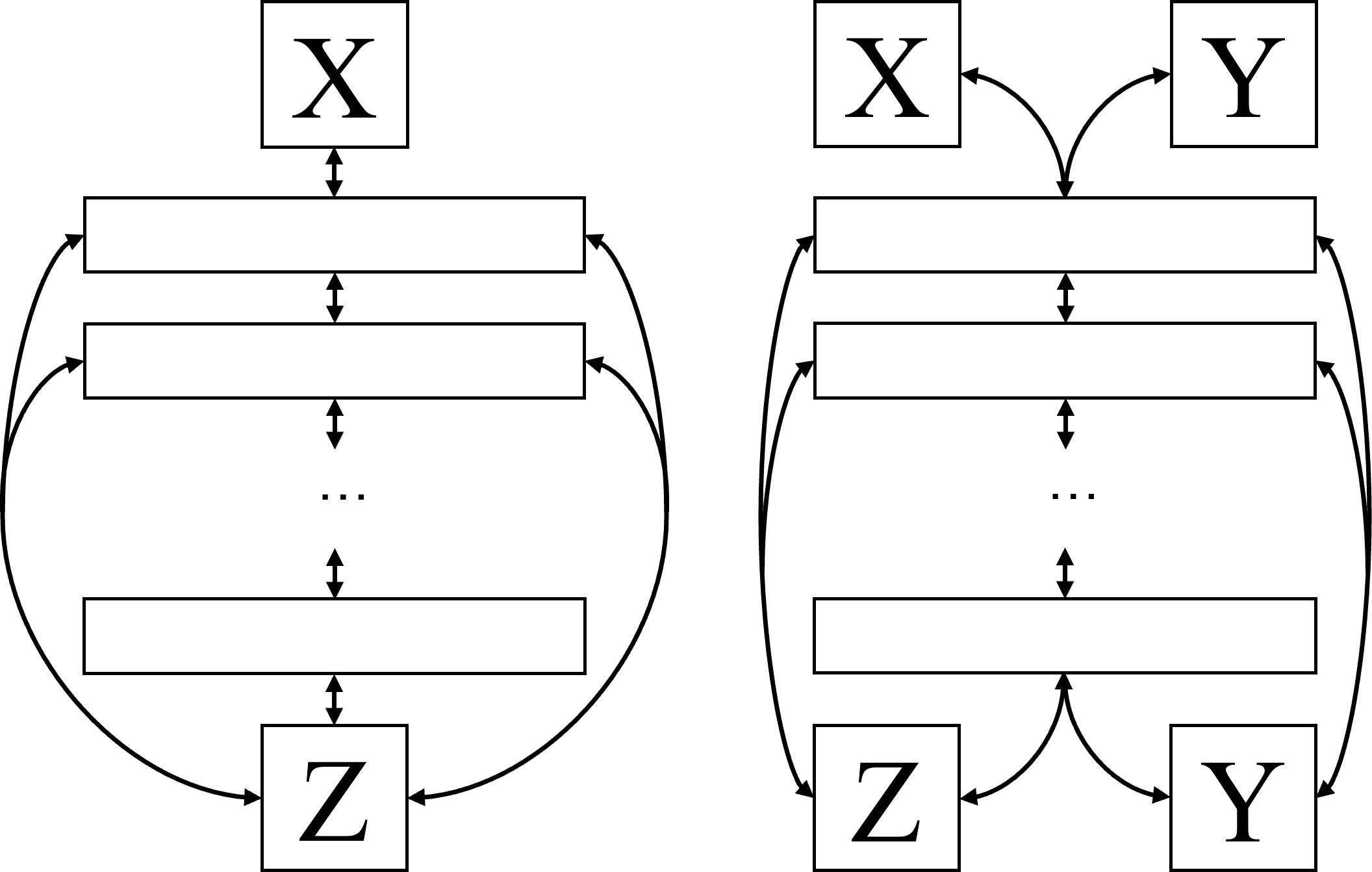}
\caption{Schematic of the conditional flow model described in this paper (right), architecturally similar to a non-conditional Real NVP flow model (left) except for the presence of $Y$ in the inputs and outputs. Arrows on the outside of the figure represent squeeze/factor operations.} 
\label{fig:schematic}
\end{figure}

$\mathbf{y}$ is now both an input and an output, so it is \textit{implicitly} embedded into every layer in $F$. Uniquely, $F$'s architecture is almost identical to a non-conditional flow, as shown in Fig.\ \ref{fig:schematic}.

\subsection{Training in Two Stages}
\label{subsec:stages_of_training}

Training minimizes Eq.\ \ref{eq:final_training_objective} using data examples drawn from the joint data distribution, $(\mathbf{x},\mathbf{y}) \sim p^\textrm{data}_{XY}(\mathbf{x},\mathbf{y})$. I observe that if $\lambda$ is set large enough that the second term dominates early on, training occurs in two distinct stages.

During the first stage, the model trains to both preserve the conditioning input $\mathbf{y}$ throughout the map, and satisfy maximum likelihood under $p_{Y'}$. By the time the other two terms begin contributing significantly to the loss, $d(F_{Y'}(\mathbf{x},\mathbf{y}),\mathbf{y}) \approx 0$. In my experiments, this stage usually finishes before the end of the first epoch.

The second stage occupies the rest of training and proceeds almost identically to the non-conditional case, except that $F$ now depends on both $\mathbf{x}$ and $\mathbf{y}$. To see this, compare the first and third terms in Eq.\ \ref{eq:final_training_objective} to Eq.\ \ref{eq:log_change_of_variables_XZ} and write the Jacobian in block matrix form:
\begin{equation}
\label{eq:Jacobian_of_fxy}
\nonumber
    \mathbb{J}_F(\mathbf{x},\mathbf{y})
    =
    \frac
        {\partial F(\mathbf{x},\mathbf{y})}
        {\partial [\mathbf{x},\mathbf{y}]^\mathsf{T}}
    =
    \left[
    \begin{array}{cc}
        \frac
        {\partial F_Z(\mathbf{x},\mathbf{y})}
        {\partial \mathbf{x}^\mathsf{T}}
        &
        \frac
        {\partial F_Z(\mathbf{x},\mathbf{y})}
        {\partial \mathbf{y}^\mathsf{T}}
         \\
         \\
        \frac
        {\partial F_{Y'}(\mathbf{x},\mathbf{y})}
        {\partial \mathbf{x}^\mathsf{T}}
        & 
        \frac
        {\partial F_{Y'}(\mathbf{x},\mathbf{y})}
        {\partial \mathbf{y}^\mathsf{T}}
    \end{array}
    \right].
\end{equation}
After the first stage, $F_{Y'}(\mathbf{x},\mathbf{y})\approx\mathbf{y}$, so the bottom right term goes to $\mathbb{1}$. Similarly, $Z$ and $Y'$ are independent, so $F_Z(\mathbf{x},\mathbf{y})$ is independent from $\mathbf{y}$ and the top right term goes to $\mathbb{0}$. Then,
\begin{equation}
\label{eq:breaking_up_J}
\nonumber
    \det
    \mathbb{J}_F(\mathbf{x},\mathbf{y})
    \approx
    \det
    \left(
    \partial F_Z(\mathbf{x},\mathbf{y})
    /
    \partial\mathbf{x}^\mathsf{T}
    \right),
\end{equation}
which closely resembles the Jacobian term in Eq.\ \ref{eq:log_change_of_variables_XZ}.

\subsection{Factoring Out the Condition Distributions}
\label{subsec:conditional_distributions}

The second stage is equivalent to factoring $p_Y$ out from both sides of Eq.\ \ref{eq:change_of_variables_XYZY} and training on maximum likelihood under the conditional distribution
\begin{equation}
\label{eq:final_training_objective_variant_2}
    p_{X\vert Y=\mathbf{y}}(\mathbf{x})
    \\
    =
    p_Z(F_Z(\mathbf{x},\mathbf{y}))
    \left\lvert
    \det \left(
        \partial 
        F_Z(\mathbf{x},\mathbf{y})
        /
        \partial
        \mathbf{x}^\mathsf{T}
    \right)
    \right\rvert.
\end{equation}
For discrete $Y$, if all conditional distributions $p^\textrm{data}_{X\vert Y=\mathbf{y}}(\mathbf{x})$ are well-represented in the training data, a model might bypass the first stage and directly optimize Eq.\ \ref{eq:final_training_objective_variant_2}. If $Y$ is continuous and available data consists only of isolated $(\mathbf{x},\mathbf{y})$ pairs, this is not possible. Training \textit{must} be performed using the joint distribution and Eq.\ \ref{eq:final_training_objective} to reach the second stage.

This is distinct from the non-invertible conditional flow models discussed in Section\ \ref{subsec:cond_flow_models}. Here, $F$ implicitly optimizes $p_Y$ during the first stage, does not require additional loss terms, and $\mathbf{y}$ enters \textit{invertibly} into $F$. Eq.\ \ref{eq:final_training_objective_variant_2} is more similar to the loss function used to train the conditional flow model of \citeauthor{cINNs_7} (\citeyear{cINNs_7}), with the tradeoff being that instead of effectively two-stage training, their design parameterizes $F$ on $\mathbf{y}$ using a two-stage model. At the same time, the implicit condition-dependent parameterization of each layer here is learned during training, rather than explicit layer-to-layer coupling as in their two-step process.

\section{Implementation Details}
\label{sec:model_architecture}

My experiments use a modified implementation of Real NVP \citep{RealNVP}, although the concept is equally applicable to other normalizing flows. The documented code used in this paper is available at \text{https://github.com/USArmyResearchLab/ARL\_} \text{Conditional\_Normalizing\_Flows}. Further experimental details are available in the technical appendix.

\subsubsection{Model Architecture}
\label{subsubsec:model_architecture}

I use a standard normal Gaussian prior, $p_Z(\mathbf{z})\sim\mathcal{N}(\mathbf{0},\mathbf{1})$, and set $\lambda=100$ and $d(F_{Y'}(\mathbf{x},\mathbf{y}),\mathbf{y})=\lVert F_{Y'}(\mathbf{x},\mathbf{y})-\mathbf{y}\rVert_1$ in Eq.\ \ref{eq:final_training_objective}. $F$ is a Real NVP model built from dense neural networks for the toy problems and multi-scale, ResNeXt-based convolutional neural networks \citep{ResNet, ResNeXt} for the image problems, incorporating identity maps \citep{ResNet_identity} and dilated convolution kernels. I use layer normalization \citep{Layer_Normalization} instead of batch normalization and drop the redundant $L^2$ weight normalization \citep{L2_weights_vs_batch_layer_norm}. I reshape layer inputs to remove masked elements, rather than replace them with zeros. Squeeze/factor operations pass elements to both the $Z$  and $Y$ components of the output.

\subsubsection{Ensuring $F(\mathcal{X},\mathcal{Y})$ Is Supported On $(\mathcal{Z},\mathcal{Y})$}
\label{subsubsec:overlapping_support}

I train $F$ in three steps. First, pre-condition the model on the latent prior, with $\mathbf{x}\sim p_Z(\mathbf{x})$ and $\mathbf{y}\sim p_Z(\mathbf{y})$, until it converges. Then, anneal in the training and validation data over $100$ epochs. Finally, train the model normally until it converges.

\subsubsection{Avoiding Latent Space Sorting}
\label{subsubsec:latent_sorting}
During training, Eq.\ \ref{eq:final_training_objective} is evaluated as an average over the batch. For discrete (class) conditions, to guarantee that $F$ does not internally sort each class into a different region of latent space, I segregate the training data by batch: each batch contains examples from only one class. I do not do this for continuous conditions.

\subsubsection{Matching Dimensionality Between $\mathcal{Z}$ and $\mathcal{X}$}
\label{subsubsec:dimension_mismatch_MNIST}
Images in MNIST and fashion-MNIST have an irregularly shaped, zero-valued background that makes up much of the total image area. $F$ cannot bijectively map the ``distribution" of observed values in a background pixel (a delta function) to the latent prior. Similarly, due to the 8-bit compression of the images, foreground pixels are represented by a comb of delta functions corresponding to the discrete intensity values.

The discrete data can be dequantized by replacing each discrete intensity value with a uniform distribution \citep{dequantization_2,dequantization}. However, the uniform distribution is still bounded, unlike the Gaussian prior, and is therefore not a natural fit for the output of a flow model \citep{INNs_flowplusplus}. I instead perform a rough unbounded dequantization on the data, $(\mathbf{x},\mathbf{y})\leftarrow (1-\alpha)(\mathbf{x},\mathbf{y}) + \alpha\mathcal{N}(\mathbf{0},\mathbf{1})$, with $\alpha=0.02$, regenerating the noise every epoch. This is enough to confuse nearby pixel intensity differences, but not semantic content.

\section{Experiments and Discussion}
\label{sec:experimental_results}

My experiments fall into two sets of overlapping categories: toy vs. image problems, and discrete vs. continuous problems. The toy problems are all 3-dimensional, with $\mathbf{x}=(x_1,x_2)$ the Cartesian coordinates of a point and $y$ the corresponding scalar condition, and serve to illustrate model properties. The image problems primarily use fashion-MNIST \citep{fMNIST}, with MNIST \citep{MNIST} included for comparison, where the image $\mathbf{x}$ and condition $\mathbf{y}$ have the same spatial shape.

\subsection{Discrete Toy Problem: Interleaved Crescents}
\label{subsec:discrete_toy_problem}
\begin{figure}[t]
\centering
\includegraphics[width=1.0 \columnwidth]{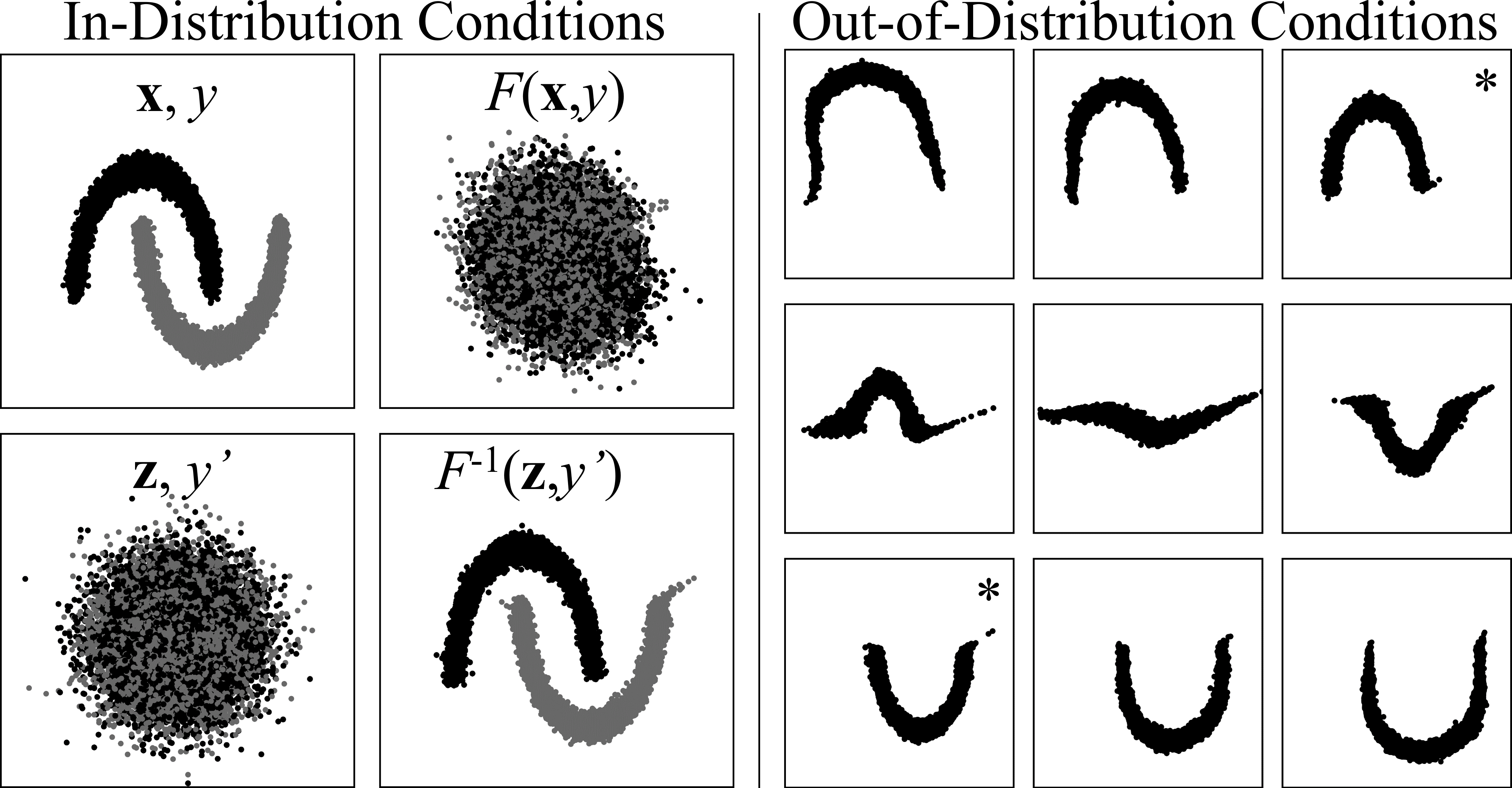}
\caption{Toy discrete problem. Left side: Training (top row) and inference (bottom row) for a flow model $F$ conditioned on class label $y=\pm 1$. Right side: Inference on $F$, with $y\in[-2,2]$ in increments of $0.5$ ($y=\pm 1$ are starred). The predictions degrade gracefully for out-of-distribution $y$.} 
\label{fig:crescents_combined}
\end{figure}

Each point $\mathbf{x}$ in these experiments is randomly sampled from one of two interleaved half-circles with finite thickness, each associated with a class label $y$. As shown on the left side of Fig.\ \ref{fig:crescents_combined}, each crescent independently maps to and from the prior. The model performs equally well if the classes overlap or they have different shapes, sizes, and topologies (an example is shown in the technical appendix). It is also robust to out-of-distribution inputs, such as the different $y$ shown on the right side of Fig.\ \ref{fig:crescents_combined}. Interpolating $y$ between class labels transitions smoothly between them, and the model preserves inputs $F_Y^{-1}(\mathbf{z},y)=y$ (not shown) and qualitative features, even when $y$ is well outside the expected domain.

\begin{figure}[t]
\centering
\includegraphics[width=0.75 \columnwidth]{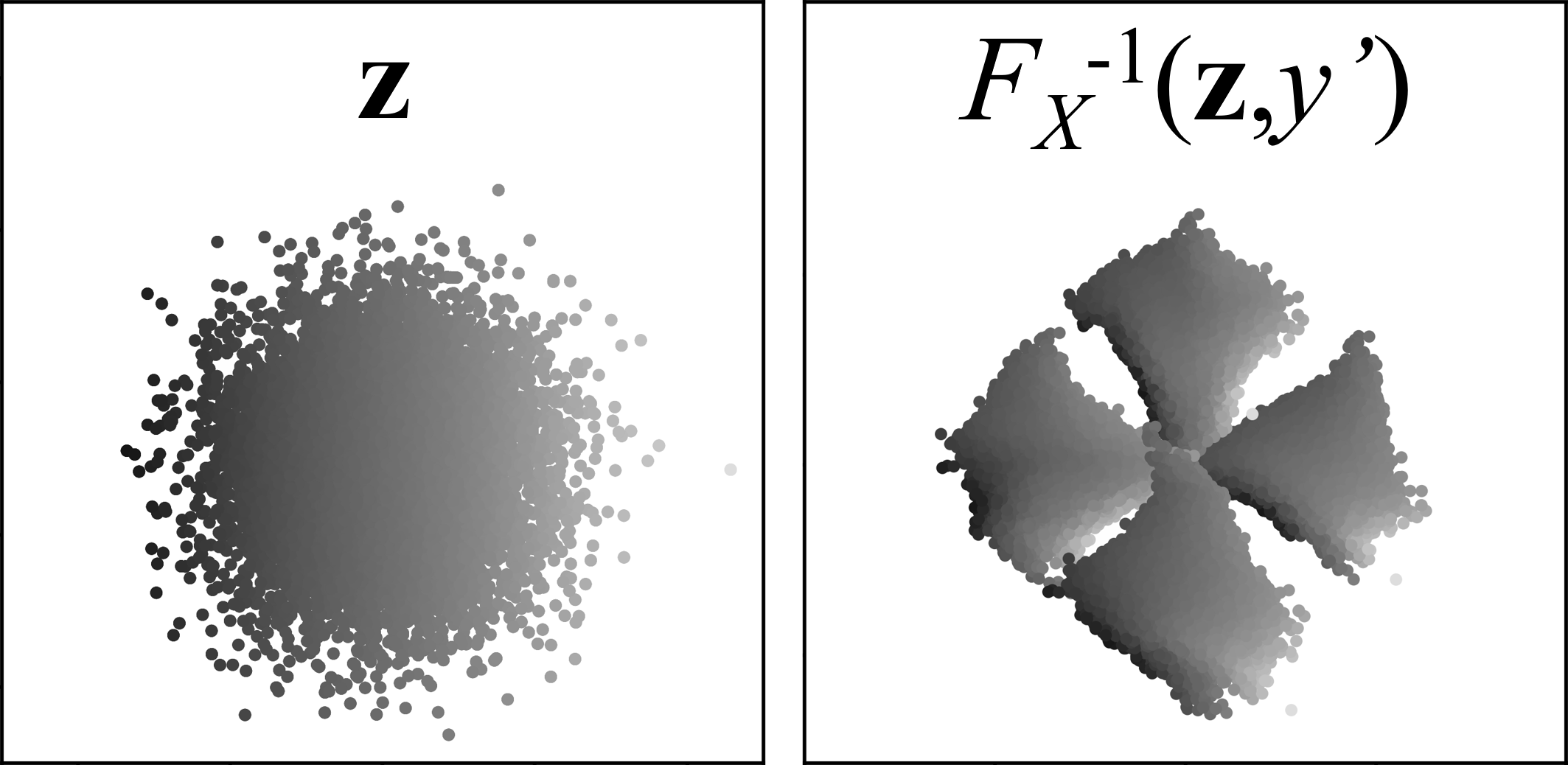}
\caption{Inference for toy continuous problem, showing $\mathbf{z}=(z_1,z_2)$ and conditional distributions for $y'\in [0,\pi/2,\pi,3\pi/2]$. In both panels, each point is shaded proportional to $z_1$. The conditional distributions all roughly preserve the spatial orientation of the original $p_Z$.} 
\label{fig:cont_cond_sec_dir}
\end{figure}

\subsection{Continuous Toy Problem: Randomly Oriented Unit Circle Sectors}
\label{subsec:continuous_toy_problem}

The conditional distributions in this case are randomly oriented sectors of the unit circle. To generate each $(\mathbf{x},y)$, I first sample an angle $y$ from the uniform distribution $\mathcal{U}_{[0,2\pi)}$, and randomly select one point $\mathbf{x}$ from the sector of width 1 radian centered on $y$. I train the model on the full joint distribution; during inference, specifying the angle yields a particular conditional distribution, as shown in Fig.\ \ref{fig:cont_cond_sec_dir}. The model obviously captures the conditional distributions correctly, but not the underlying structure of the problem.



This is because Real NVP is limited to scaling and translation transformations, while the problem has rotational symmetry. $F$ stretches and contracts the coordinate plane, parameterized on $y$. This produces an ``unnatural" map, as shown by the gradients in the figure: $F$ is one of infinitely many ``valid" maps obtained from training with incorrect inductive bias, as discussed in Section\ \ref{subsec:prob_generative_model}.

\subsection{Class-Conditional Image Generation}
\label{subsec:discrete_real_problem}
Here, each $\mathbf{x}$ is a $28\times 28$ grayscale image with discrete class label $y$. $\mathbf{y}$ is a tensor of the same spatial shape as $\mathbf{x}$, where every element $y_{ij}=y$. $(\mathbf{x},\mathbf{y})$ is standardized to mean $0$, standard deviation $1$ across the entire dataset.

\begin{figure}[t]
\centering
\includegraphics[width=1.0 \columnwidth]{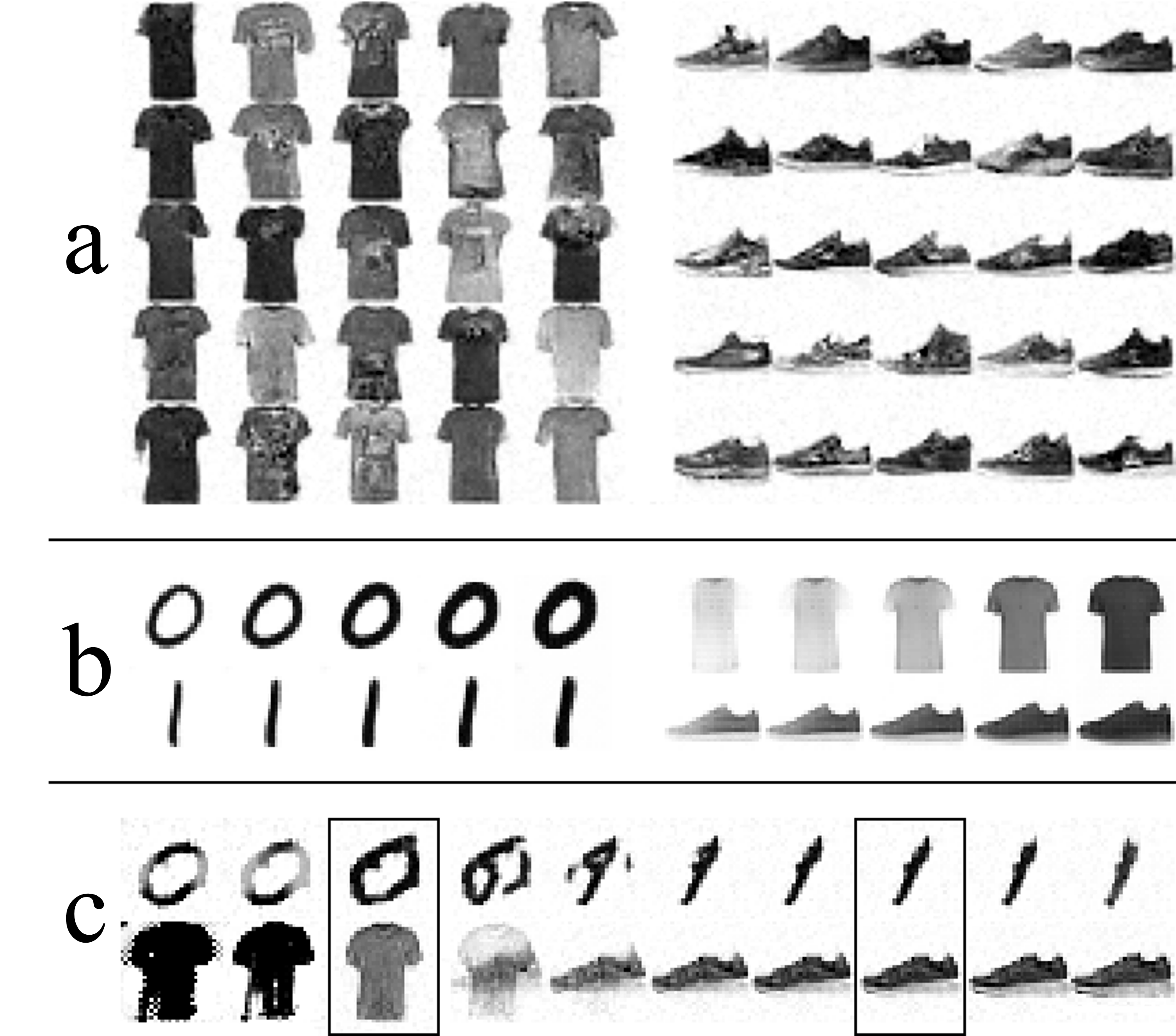}
\caption{Class-conditioned model predictions. (a) 25 randomly generated examples each for shirts and sneakers. (b) Interpolations along the same line in $\mathcal{Z}$ for different $\mathbf{y}$. (c) Interpolations along a line in $\mathcal{Y}$ connecting two classes for fixed $\mathbf{z}$ (predictions on the true class labels are boxed).}
\label{fig:class_generative}
\end{figure}

Fig.\ \ref{fig:class_generative}a shows samples generated from two learned conditional distributions with labels ``shirt" and ``sneaker." Qualitatively, the images have similar realism and diversity of features as the original dataset. The model also maintains semantic consistency in the latent space across classes (shown in Fig.\ \ref{fig:class_generative}b), and degrades gracefully for out-of-distribution $y$ (Fig.\ \ref{fig:class_generative}c). As with the toy problem, $F$ not only interpolates between true class labels $y_1$ and $y_2$ (the boxed images), it exaggerates class features for $y$ along the same line.

\subsection{Super-Resolution of Highly Degraded Images}
\label{subsec:continuous_real_problem}
\begin{figure}[t]
\centering
\includegraphics[width=1.00 \columnwidth]{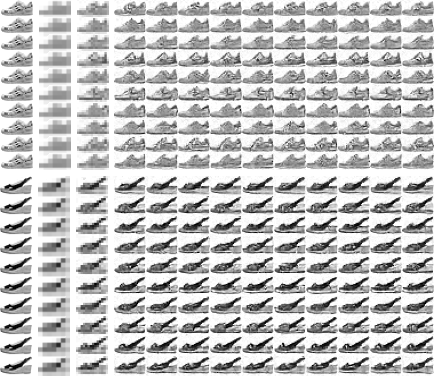}
\caption{Two-scale super-resolution of fashion-MNIST sneakers (top) and sandals (bottom), one example each. Column 1: ground truth. Column 2: $4\times 4$-downsampled model input. Column 3: 10 different intermediate-resolution reconstructions. Columns 4-13: 10 different high-resolution reconstructions for each intermediate reconstruction in column 3. Image backgrounds have been cropped.}
\label{fig:fMNIST_SR41}
\end{figure}

Here, $F$ reconstructs highly degraded low-resolution images. I spatially downsample the original $28\times 28$ images using $4\times 4$ average pooling to obtain $7\times 7$ inputs. As shown in the first and second columns of Fig.\ \ref{fig:fMNIST_SR41}, this destroys all but the highest-level features in the original image.

I factor reconstruction into two steps, with two composed models $F=F_1\circ F_2$. $F_2$ samples from the conditional distribution of possible $14\times 14$ intermediate images consistent with a given $7\times 7$ input condition. Then, given a sample from that intermediate distribution, $F_1$ samples from the conditional distribution of possible consistent $28\times 28$ images.

Both models take their $\mathbf{x}$ to be the residual image at that resolution, while $\mathbf{y}$ is spatially upscaled to the same size as $\mathbf{x}$. This ensures that $\mathbf{x}$ and $\mathbf{z}$ contain the same information and do not contain $\mathbf{y}$, which is not the case if $\mathbf{x}$ can simply be downsampled to get $\mathbf{y}$. Then $\mathbf{x}+\mathbf{y}$ gives the predicted image at that resolution. I train the models on the joint distribution $p_{XY}^\textrm{data}(\mathbf{x},\mathbf{y})$, including all 10 classes, \textit{without} class labels.

Unlike with non-conditional flow models, where $p_X^\textrm{data}(\mathbf{x})$ is available, it is not possible to \textit{directly} verify that $F$ samples from $p_{X\vert Y=\mathbf{y}}(\mathbf{x})$, since (unlike with the toy continuous problem) I do not have access to the distribution of real images that downsample to a given low-resolution $\mathbf{y}$. However, Eq.\ \ref{eq:final_training_objective_variant_2} states that this is the case, and the toy experiments in Section\ \ref{subsec:continuous_toy_problem} and the histogram analysis in Section\ \ref{subsec:histograms} provide strong supporting evidence.

Fig.\ \ref{fig:fMNIST_SR41} shows the results of this procedure for one example each of fashion-MNIST sneakers and sandals; the model performs equally well on the other 8 classes. The first column represents the ground truth, and the second, the $4\times 4$-downsampled low-resolution $\mathbf{y}$. Note that, after downsampling, almost all features have been lost.

The third column shows 10 predictions from $F_2^{-1}$, which reintroduce intermediate-scale features such as overall shape and pattern. The remaining columns show 10 predictions from $F_1^{-1}$ for each image in column 3, showing variation in smaller details. Each reconstruction can be downsampled to recover its low-resolution condition.

The reconstructions are sometimes ambiguous. For example, although most sandal reconstructions in Fig.\ \ref{fig:fMNIST_SR41} resemble the original, some resemble sneakers shaded dark at the front and light at the back, a pattern that occurs occasionally in the training data. This ambiguity is good! The downsampling process destroys most of the information in the original image, so reconstructions should properly reflect that uncertainty. The symmetric and invertible nature of flow models makes it easy to quantitatively verify this as well.

\subsection{Quantitative Evaluation of Model Quality}
\label{subsec:histograms}

\begin{figure}[t]
\centering
\includegraphics[width=1.0 \columnwidth]{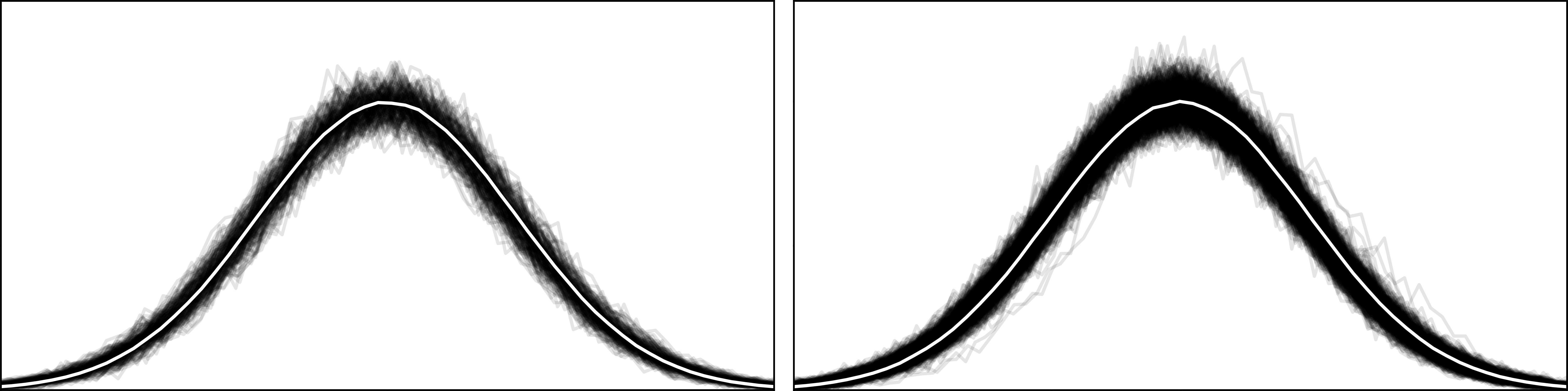}
\caption{Histograms for evaluating how well the model maps validation examples to the latent prior for fashion-MNIST super-resolution step 1 (left side) and step 2 (right side). Each black line corresponds to one pixel in the input image domain. White lines show the prior.}
\label{fig:histograms}
\end{figure}

One way to quantify the ``trustworthiness" of $F$ is to evaluate the properties of the distributions obtained from validation data. The thickness of the envelope of all distributions (shown for both super-resolution steps in Fig.\ \ref{fig:histograms}) gives a measure of how well the model maps $p_{X\vert Y=\mathbf{y}}(\mathbf{x})$ to $p_Z(\mathbf{z})$, and vice versa. If each element in $F_Z(\mathbf{x},\mathbf{y})$ approximates the prior well over $(\mathcal{X},\mathcal{Y})$, the model is sampling from the right data distribution (although $F$ may not be a good map; see Sections\ \ref{subsec:prob_generative_model} and\ \ref{subsec:continuous_toy_problem}). To my knowledge, such simple tests have not been performed in the literature on conditional models, and are only very rarely performed on non-conditional models, usually averaged over all latent dimensions (see, for example, Papamakarios et al., \citeyear{MAFs}, Nalisnick et al., \citeyear{flow_for_OOD_1}, and Kirichenko et al., \citeyear{flow_for_OOD_2}). In my experiments, I occasionally find that a small number of element histograms deviate significantly from the prior over $(\mathcal{X},\mathcal{Y})$ (an example is shown in the technical appendix), suggesting that this averaging overlooks some types of error.

Each black line in the figure corresponds to the statistics of a single pixel over the entire validation set, while the white line shows the prior, $p_Z(\mathbf{z})=\mathcal{N}(\mathbf{0},\mathbf{1})$. Together, all these histograms fit the prior well, with a fuzzy envelope. Counterintuitively, the super-resolution models have about half the envelope thickness of the class-conditional generative models (shown in the technical appendix), possibly because $p_{X\vert Y=\mathbf{y}}(\mathbf{x})$ is narrower when $\mathbf{y}$ is a low-resolution image rather than a class label. This also validates the two-step approach to training I take for the super-resolution models in Section\ \ref{subsec:continuous_real_problem}.

Analysis of a flow model's latent space in the literature is mostly limited to feature algebra. Evaluation of $F_X^{-1}(\mathbf{z})$ usually focuses on qualitative properties like diversity and realism, or computationally expensive quantitative metrics like the Fr\'{e}chet Inception Distance \citep{FID}. The latent space also provides a natural way to independently quantify model quality, for negligible cost, that can also identify potential problem areas during or after training.

\section{Conclusion}
\label{sec:conclusion}

The conditional flow model described in this paper is a modular extension to standard non-conditional Real NVP, meaning it should be possible to directly apply recent advances in flow model design such as learned masks \citep{INNs_glow}, improved inductive biases \citep{flow_for_OOD_2}, or a coupling law that is more expressive \citep{NAFs} or incorporates self-attention \citep{INNs_flowplusplus}. A key part of future work is designing a model tailored to the specific problem at hand.

To avoid having to model $p_Y(\mathbf{y})$ directly, the condition variable must be the same for both the input and output (i.e., $Y=Y'$), meaning this method is unsuitable for problems like domain translation. Otherwise, it provides a simple and inexpensive way to solve ill-posed problems via density modeling. At least in principle, the method provides direct access to the detailed statistics of the predicted distributions; further, the predictions appear robust to out-of-distribution conditioning inputs. Although more work remains to be done, this is a promising, low-cost approach to rigorously solving ill-posed problems.

\section*{Acknowledgements}
I would like to thank Dr. Michael S. Lee for helpful suggestions. Compute time was provided by the DEVCOM ARL DoD Supercomputing Resource Center.

\bibliography{bibliography}

\end{document}


\maketitle

\appendix
\section*{A. Model Architecture}
\label{app:model_architecture}

\setcounter{equation}{0}
\renewcommand{\theequation}{A.\arabic{equation}}

\setcounter{figure}{0}
\renewcommand{\thefigure}{A.\arabic{figure}}

This Appendix describes the architecture of the models I use in my experiments in more detail.

\subsection*{Real NVP Architecture}
\label{subsec:flow_architecture}

A non-conditional normalizing flow $F : X \rightarrow Z$ is defined as the composition of $N$ simpler bijective transformations, $F = f_1 \circ f_2 \circ \cdots \circ f_N$. Those transformations map intermediate states, $f_i : U_i \rightarrow V_i$, where $U_1 = X$, $V_i = U_{i+1}$, and $V_N = Z$. Each $\mathbf{u}_i$ and $\mathbf{v}_i$ are broken into two blocks:
\begin{subequations}
\begin{align}
\label{eq:defining_f_2a}
    \mathbf{v}_{i,1}
    =
    f_{i,1}(\mathbf{u}_i)
    &=
    \mathbf{u}_{i,1}
\\
\label{eq:defining_f_2b}
    \mathbf{v}_{i,2}
    =
    f_{i,2}(\mathbf{u}_i)
    &=
    f_{i,2}(\mathbf{u}_{i,2};
        \phi(\mathbf{u}_{i,1})).
\end{align}
\end{subequations}
The first block, $\mathbf{u}_{i,1}$, is unchanged as it passes through $f_i$. $f_{i,2}$ defines a \textit{coupling law,} which is an invertible map with respect to its first argument, given the second. $\phi$ is the \textit{coupling function,} which can be arbitrarily complex and does not have to be invertible, and is typically defined by one or more neural networks.

The Jacobian of $F$ is $\lvert\det\mathbb{J}_F\rvert=\prod_i\lvert\det\mathbb{J}_{f_i}\rvert$. For an affine coupling law,
\begin{subequations}
\begin{align}
\nonumber
    f_{i,1}(\mathbf{u}_i)
    &=
    \mathbf{u}_{i,1}
\\
\nonumber
    f_{i,2}(\mathbf{u}_i)
    &=
    \mathbb{A}_i(\mathbf{u}_{i,1})
    \mathbf{u}_{i,2}
    +
    \mathbf{b}_i(\mathbf{u}_{i,1}),
\\
\nonumber
    f_{i,1}^{-1}(\mathbf{v}_i)
    &=
    \mathbf{v}_{i,1}
\\
\nonumber
    f_{i,2}^{-1}(\mathbf{v}_i)
    &=
    \mathbb{A}^{-1}_i(\mathbf{v}_{i,1})
    (\mathbf{v}_{i,2} - \mathbf{b}_i(\mathbf{v}_{i,1})),
\end{align}
\end{subequations}
where $\mathbb{A}_i$ and $\mathbf{b}_i$ are both functions of $\mathbf{u}_{i,1}=\mathbf{v}_{i,1}$. The associated Jacobian and its determinant are then
\begin{align}
\nonumber
    \mathbb{J}_{f_i}(\mathbf{u}_i)
    &=
    \left[
    \begin{array}{cc}
        \frac
        {\partial f_{i,1}(\mathbf{u}_i)}
        {\partial \mathbf{u}_{i,1}^\mathsf{T}}
        &
        \frac
        {\partial f_{i,1}(\mathbf{u}_i)}
        {\partial \mathbf{u}_{i,2}^\mathsf{T}}
        \\
        \\
        \frac
        {\partial f_{i,2}(\mathbf{u}_i)}
        {\partial \mathbf{u}_{i,1}^\mathsf{T}}
        & 
        \frac
        {\partial f_{i,2}(\mathbf{u}_i)}
        {\partial \mathbf{u}_{i,2}^\mathsf{T}}
    \end{array}
    \right]
\\
\nonumber
    &=
    \left[
    \begin{array}{cc}
        \mathbb{1}
        &
        \mathbb{0}
        \\
        \\
        \frac
        {\partial f_{i,2}(\mathbf{u}_i)}
        {\partial \mathbf{u}_{i,1}^\mathsf{T}}
        & 
        \mathbb{A}_i
    \end{array}
    \right],
\end{align}
and $\lvert\det \mathbb{J}_{f_i}(\mathbf{u}_i)\rvert=\lvert\det \mathbb{A}_i(\mathbf{u}_{i,1})\rvert$. Real NVP \citep{RealNVP} uses scaling affine transformations only ($\mathbb{A}_i$ is diagonal), so $\lvert\det \mathbb{A}_i\rvert=\lvert\prod_j A_{i,jj}\rvert$.

The conditional flow models in this paper have identical architecture, except that they define a map $F : (X,Y) \rightarrow (Z,Y')$ with $U_1=(X,Y)$ and $V_N=(Z,Y')$. $F$ treats the condition component of the input as an additional channel concatenated to the non-condition (data or prior) component; that is, $Y$ is concatenated to $X$ and $Y'$ is concatenated to $Z$.

The flow models used for the toy problems are very simple, and do not use compressed masking (described in the next section): each ``block" of coupling layers consists of 6 masks, one for each nontrivial combination of the three input dimensions: $(1,0,0)$, $(0,1,0)$, $(0,0,1)$, $(0,1,1)$, $(1,0,1)$, and $(1,1,0)$, randomly ordered. All toy experiments in the paper are performed with 4 sets of these coupling blocks (24 coupling layers total), although I observe similar results with as few as 1 block. For the continuous toy problem described in Section 6.2, even with as many as 20 coupling blocks (120 coupling layers total), as expected, the model is still unable to ``learn rotations" with a scaling coupling law.

For the toy problems, I also experimented with affine transformations that include shear. In this case, $\mathbb{A}_i$ is triangular instead of diagonal, so the Jacobian of $F$ is unchanged. In my experiments, this leads to more unstable training, without any noticeable increase in performance. This is probably because for the datasets in my experiments, none of them can be mapped to the prior with scaling and shear transformations much more effectively than with scaling alone.

The flow models used for the image problems use coupling blocks with 4 coupling layers each, as there are two complementary checkerboard masks and two complementary channelwise masks. They do use compressed masking (described in the next section). All image experiments in the paper use 4 sets of these coupling blocks (16 coupling layers total), although I observe similar results with between 3 and 6 blocks.

As in the original Real NVP design, for the image problems, I use squeeze/factor operations to factor out elements to pass directly to the output after some layers. Equal numbers of elements are passed to $Z$ and $Y'$ per squeeze/factor operation, rather than just $Z$, as shown in Fig.\ 2. I observe similar results with or without these squeeze/factor operations (the latter is possible because of the small size of fashion-MNIST and MNIST).

\subsection*{Compressed Masking}
\label{subsec:compressed_masking}
In the original Real NVP model, $\mathbf{u}_{i,1}$ and $\mathbf{u}_{i,2}$ are derived from $\mathbf{u}_i$ via masking, using alternating checkerboard and channelwise masks in sets of four. The masked elements are first set to zero. Then, after passing $\mathbf{u}_{i,2}$ through $f_{i,2}$, the mask is re-applied (since those elements may be nonzero now) to obtain $\mathbf{v}_{i,2}$, and the result added to $\mathbf{v}_{i,1}=\mathbf{u}_{i,1}$ to obtain $\mathbf{v}_i$. The same process is applied in reverse for $f_{i,2}^{-1}$.

\begin{figure}[t]
\centering
\includegraphics[width=0.75\columnwidth]{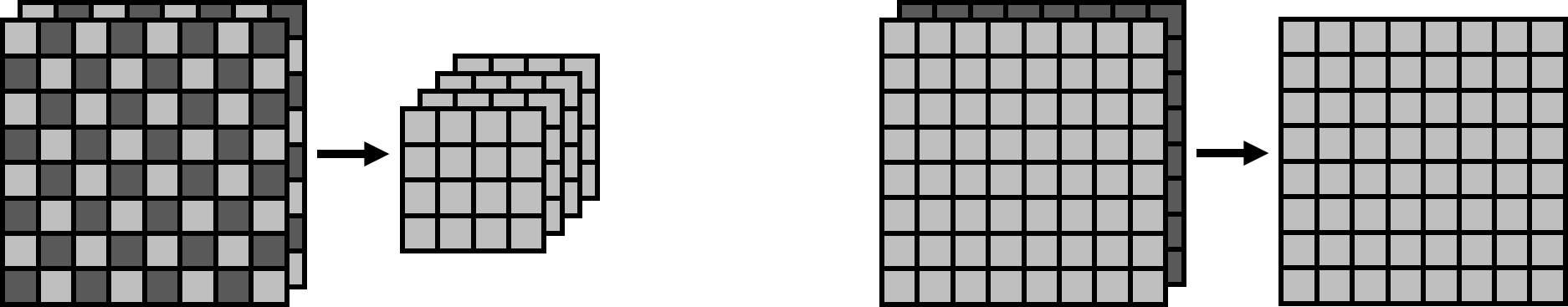}
\caption{The compressed masking procedure for checkerboard masks (left) and channelwise masks (right). Dark squares are masked pixels, which are compressed out before being passed into the coupling layer.} 
\label{fig:masking}
\end{figure}

Rather than pass extra masked (zeroed) elements through $f_{i,2}$ and re-masking them afterward, I compress my inputs by reshaping $\mathbf{u}_i$ to remove the masked elements. With 1-dimensional vector inputs $\mathbf{u}_i=[\mathbf{u}_{i,1},\mathbf{u}_{i,2}]$, it is possible to define rectangular mask matrices $\mathbb{M}_1$ and $\mathbb{M}_2$ such that
\begin{subequations}
\begin{align}
    \nonumber
    \mathbb{M}_1\mathbf{u}_i
    &=\mathbf{u}_{i,1}
    \\
    \nonumber
    \mathbb{M}_2\mathbf{u}_i
    &=\mathbf{u}_{i,2}
    \\
    \nonumber
    \mathbf{v}_{i,1}\mathbb{M}_1^\mathsf{T}
    &=[\mathbf{v}_{i,1},0]
    \\
    \nonumber
    \mathbf{v}_{i,2}\mathbb{M}_2^\mathsf{T}
    &=[0,\mathbf{v}_{i,2}].
\end{align}
\end{subequations}
The relevant components of $\mathbf{u}_i$ are reduced in size using $\mathbb{M}_j\mathbf{u}_i$, passed into Eqs.\ \ref{eq:defining_f_2a} and \ref{eq:defining_f_2b}, and then projected back into the original space using $\mathbf{v}_{i,j}\mathbb{M}_j^\mathsf{T}$. Analogous transformations can be defined for the checkerboard and channelwise masks used in Real NVP with image-shaped inputs, as shown in Fig.\ \ref{fig:masking}. This reduces the total input size to each neural network by a factor of 2: for checkerboard-masked inputs, both spatial dimensions are halved and the channel depth is doubled; for channelwise-masked inputs, the channel depth is halved.

\subsection*{Neural Network Architecture}
\label{subsec:NN_architecture}

For the toy problems, I define $\mathbb{A}_i$ and $\mathbf{b}_i$ using simple dense neural networks with 12 layers of 64 nodes each. I observe similar results with between 6 and 30 layers, and up to 512 nodes.

\begin{figure}[t]
\centering
\includegraphics[width=0.3625\columnwidth]{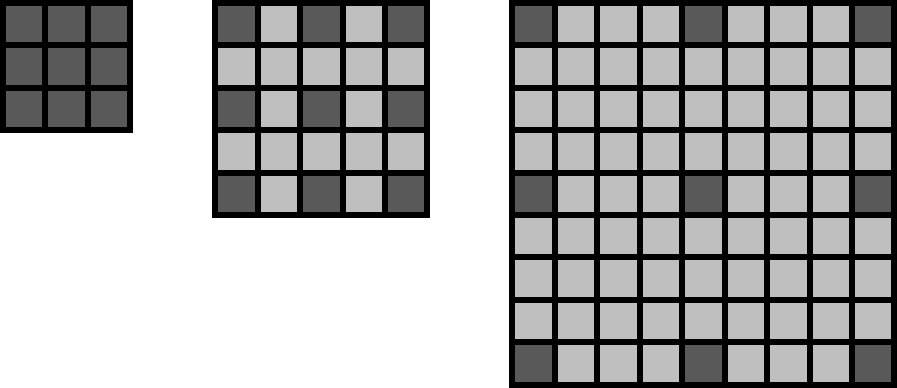}
\caption{An example of $3\times 3$ kernels used in the convolutional model, with dilation factors 1, 2, and 4. Each dilated kernel spans two adjacent pixels in the next larger dilation.} 
\label{fig:dilated_kernels}
\end{figure}

For the image problems, I define $\mathbb{A}_i$ and $\mathbf{b}_i$ using ResNeXt-based convolutional networks \citep{ResNet, ResNeXt} with identity maps \citep{ResNet_identity}. Each ResNeXt block contained a branch of $3\times 3$ convolutional kernels, separated into a number of sub-branches equal to the cardinality of the network; the experiments in the paper use 64 kernels and a cardinality of 8. Parallel to this branch, there are additional branches of dilated $3\times 3$ kernels, with the number of kernels (and cardinality) scaled down by the dilation factor of each dilated branch; the rationale is that fewer kernels should be required to learn long-range pixel correlations in MNIST and fashion-MNIST. The dilations are chosen as shown in Fig.\ \ref{fig:dilated_kernels}, with each kernel spanning 2 adjacent pixels in the next-larger dilation. Larger dilations are added following this pattern until the next-larger dilation would be larger than half the spatial size of the input image. This means that for checkerboard-masked coupling layers, there will be one fewer dilation (but because the checkerboard-masked layers are halved spatially during the compression, the total receptive field will be similar).

For the image problems I use Leaky ReLU activations followed by layer normalization \citep{Layer_Normalization}, except that the final layers used a $\tanh$ activation multiplied by a learned scale factor in $\mathbb{A}_i$ and a linear activation in $\mathbf{b}_i$. For the toy problems I omit the layer normalization. To keep the first term of Eq.\ 8 from underflowing, I initialize $\mathbb{A}_i$ and $\mathbf{b}_i$ with an orthogonal kernel initializer with gain of $0.1$.

\subsection*{Logit vs. No Logit}
\label{subsec:logit}

\begin{figure}[t]
\centering
\includegraphics[width=1.0\columnwidth]{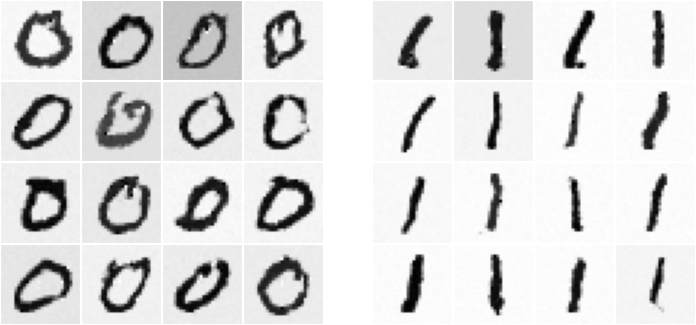}
\caption{Generated digits from a class-conditional model trained on MNIST, conditioned on labels 0 (left) and 1 (right). Note that a small number of predicted pixels have unnaturally high or low values (particularly visible in row 1, column 3 of the 0-labeled images, where a single pixel is predicted to have higher intensity than the white background).} 
\label{fig:MNIST}
\end{figure}

I experimented with training on the logit of intensity for the image problems, as in the original Real NVP paper. I obtain good results, but decided to avoid doing this for the experiments in the paper. The reason is that, both with and without logits, the model sometimes returns very high or low intensities (enough so to be obviously outside the data distribution). Examples of this are shown in Fig.\ \ref{fig:MNIST}, using a class-conditioned model trained on MNIST digits. The images are scaled to have the same minimum and maximum pixel intensity; some images appear to have unusually light strokes or unusually dark backgrounds are because, without the logit, the model occasionally predicts one or two pixels to be very bright or very dark.

Implementing a logit obscures this: although the logit function ensures that only reasonable predictions are made, it hides the fact that the model itself sometimes predicts obviously out-of-distribution elements. Rather than covering up this problem with the logit, this is evidence that these models require more work before they can be reliably deployed in the real world.

\appendix
\section*{B. Hardware and Software}
\label{app:hardware_software}

The code was written using the TensorFlow library \citep{tensorflow} and requires TensorFlow version 2.7 or higher and TensorFlow Probability 0.15 or higher to run. The models are trained on NVIDIA V100 GPUs (32 GB memory) with AMD 7H12 Rome CPUs (Linux OS).

\appendix
\section*{C. More Complex Discrete Toy Problem}

\setcounter{figure}{0}
\renewcommand{\thefigure}{C.\arabic{figure}}

\begin{figure*}[ht]
\centering
\includegraphics[width=1.0\textwidth]{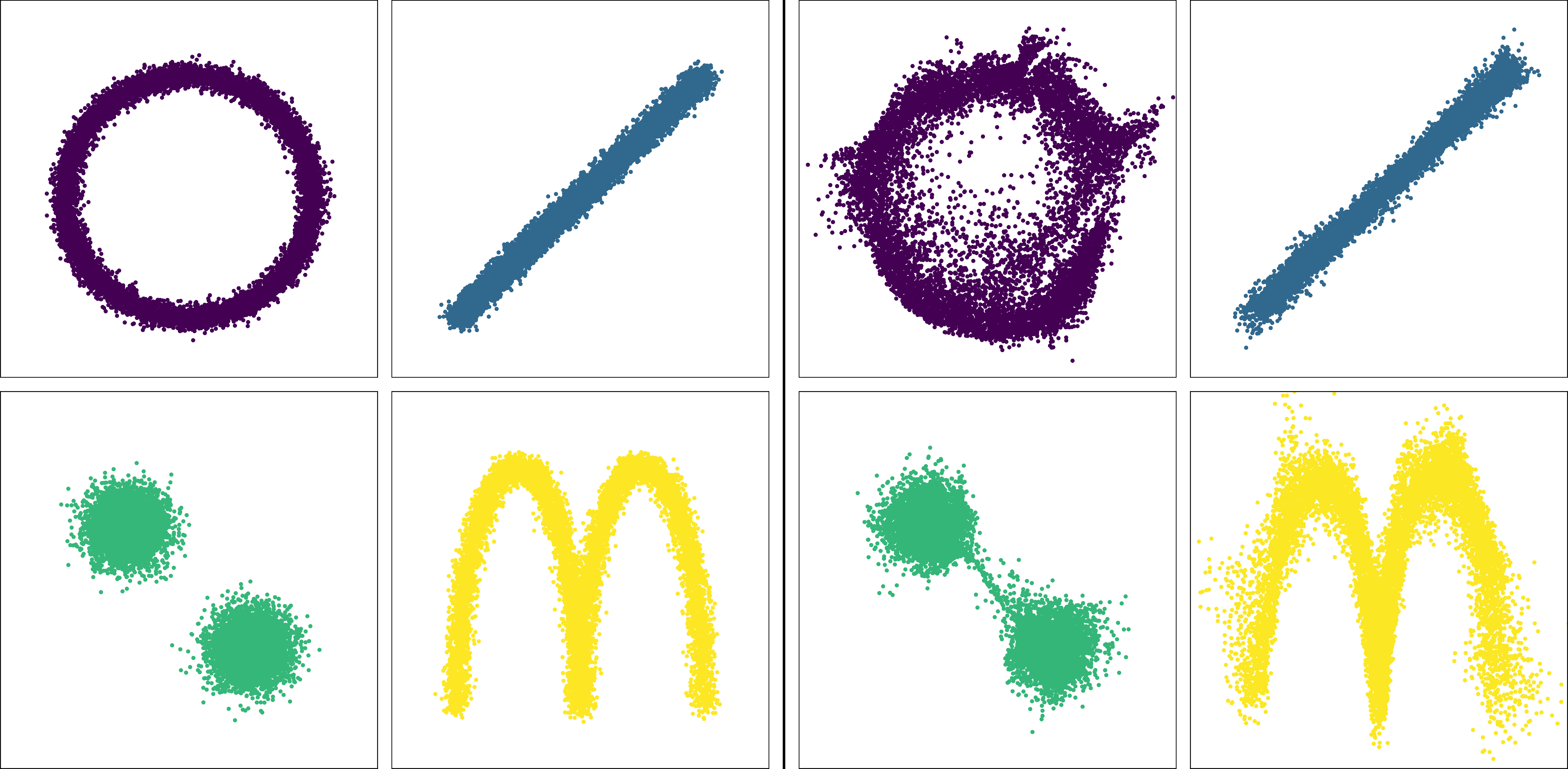}
\caption{Example of a conditional discrete problem where the target distributions overlap (so some points in different conditional distributions are distinguished only by label) and have different shapes and topologies. Left side: target data distributions. Right side: conditional distributions learned by a conditional flow model. Color represents input label on the left and output label on the right (they are identical for all mapped points).} 
\label{fig:mixed_dataset}
\end{figure*}

The conditional flow model works equally well if the classes are dissimilar or overlapping, as shown in Fig.\ \ref{fig:mixed_dataset} for a model with four labels, each corresponding to a distinct shape. Topological changes (converting the Gaussian prior to a circle or two distinct blobs) are the main source of error.

\appendix
\section*{D. Histograms of Data Mapped to $\mathcal{Z}$}
\label{app:histograms}

\setcounter{figure}{0}
\renewcommand{\thefigure}{D.\arabic{figure}}

\begin{figure}[t]
\centering
\includegraphics[width=1.0\columnwidth]{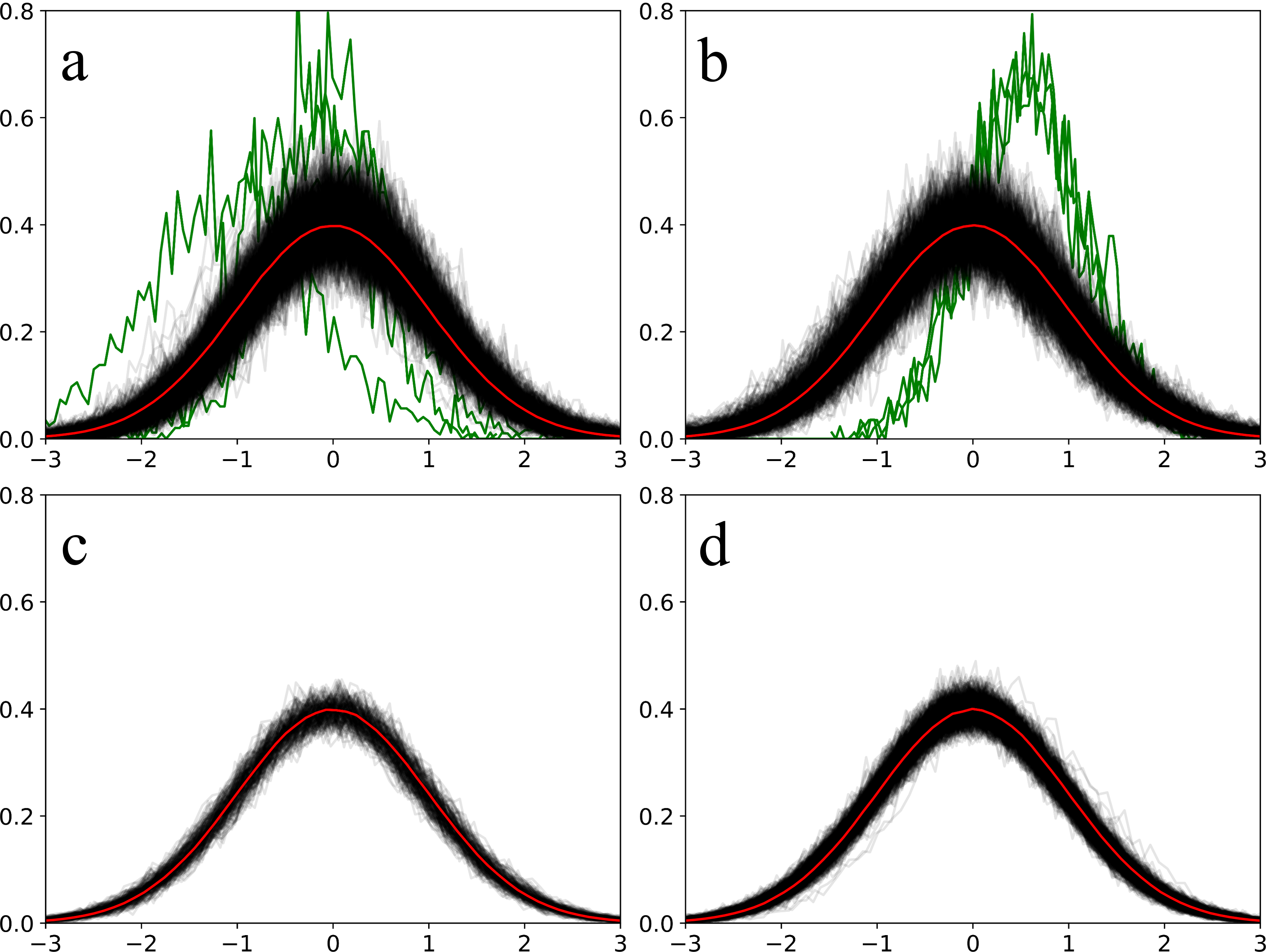}
\caption{Histograms obtained by feeding validation data through the model in the $XY\rightarrow ZY'$ direction, for class-conditional image generation on (a) MNIST and (b) fashion-MNIST, and for super-resolution on fashion-MNIST, (c) step one and (d) step two. Each pixel is represented by one black line, while the red line shows the prior. In the class-conditioned image generation, occasionally the model learns to map a small number of pixels, shown in green, to distributions significantly different from the prior.} 
\label{fig:histograms}
\end{figure}

As mentioned in the text, the histograms of data mapped to latent space (averaged over all the validation data, but \textit{not} spatially averaged over the image) for the class-conditioned models typically have an envelope about twice as wide for the class-conditioned image generators as the super-resolution models, as shown in Fig.\ \ref{fig:histograms}. Additionally, sometimes, the class-conditioned image generators map a small number of pixels to the wrong distribution. Given validation data, they map to something other than the standard normal Gaussian prior, shown by the green lines in the figure. I only observe this occasionally, and only for the class-conditioned models. In the generated data, the affected pixels map the prior to a distribution that skews lighter or darker than the surrounding pixels (and is usually in the background). Spatial averaging of the histograms loses this information, which provides a way to identify a particularly flawed map.

\appendix
\section*{E. Two-step Super-resolution Model}
\label{app:SR_algorithms}

\setcounter{algorithm}{0}
\renewcommand{\thealgorithm}{E.\arabic{algorithm}}

\setcounter{figure}{0}
\renewcommand{\thefigure}{E.\arabic{figure}}

This appendix contains the algorithms used to train and run inference on the two-step super-resolution model, and more examples from super-resolution on other classes.

\begin{algorithm}
\caption{Training the Two-step Super-resolution Model.}
\label{alg:SR_training}
\begin{algorithmic}
    \State{Original dataset contains high-resolution images $\mathbf{x}_0'$}
    \State{$\mathbf{x}'_1 \gets \downsample(\mathbf{x}'_0)$}
    \State{$\mathbf{x}'_2 \gets \downsample(\mathbf{x}'_1)$}
    \State{train $F_2$ on examples
    $(\mathbf{x}_2,\mathbf{y}_2)$
    \\ \hskip\algorithmicindent
    $= (\mathbf{x}'_1 - \upsample(\mathbf{x}'_2), \upsample(\mathbf{x}'_2))$}
    \State{train $F_1$ on examples
    $(\mathbf{x}_1,\mathbf{y}_1)$
    \\ \hskip\algorithmicindent
     $= (\mathbf{x}'_0 - \upsample(\mathbf{x}'_1), \upsample(\mathbf{x}'_1))$}
\end{algorithmic}
\end{algorithm}

\begin{algorithm}
\caption{Distribution Inference With the Two-step Super-resolution Model.}
\label{alg:SR_inference}
\begin{algorithmic}
    \State{Condition input is a single $7\times 7$ image $\mathbf{y}_2$}
    
    \State{$N_1 =$ number of $14\times 14$ reconstructions to sample}
    
    \State{$N_2 =$ number of $28\times 28$ reconstructions to sample}
    
    \State{$\mathbf{y}_2 \gets \upsample(\mathbf{y}_2)$}
    
    \For{$i$ in $[1,\dots,N_1]$}
        
        \State{$\mathbf{z}_{2,i}\sim p_{Z_2}(\mathbf{z}_2)$}
        
        \State{$\mathbf{x}_{2,i} \gets F^{-1}_{2,X}(\mathbf{z}_{2,i},\mathbf{y}_2)$} \Comment{A point sampled from
        \\ \hfill
        $p_{X_2\vert Y_2=\mathbf{y}_2}(\mathbf{x}_2)$}
        
        \State{$\mathbf{x}'_{1,i} \gets \mathbf{x}_{2,i} + \mathbf{y}_2$} \Comment{A single possible
        $14\times 14$
        \\ \hfill
        reconstruction}
        
        \State{$\mathbf{y}_{1,i} \gets \upsample(\mathbf{x}'_{1,i})$}
        
        \For{$j$ in $[1,\dots,N_2]$}
            
            \State{$\mathbf{z}_{1,j}\sim p_{Z_1}(\mathbf{z}_1)$}
            
            \State{$\mathbf{x}_{1,ij} \gets F_{1,X}^{-1}(\mathbf{z}_{1,j},\mathbf{y}_{1,i})$} \Comment{A point sampled from
            \\ \hfill
            $p_{X_1\vert Y_1=\mathbf{y}_{1,i}}(\mathbf{x}_1)$}
            
            \State{$\mathbf{x}'_{0,ij} \gets \mathbf{x}_{1,ij} + \mathbf{y}_{1,i}$} \Comment{A single possible $28\times 28$
            \\ \hfill
            reconstruction}
        
        \EndFor
    \EndFor
\State{\textbf{return} $\{\mathbf{x}_{0,ij}'\}$} \Comment{Samples from the predicted distribution
\\ \hfill
of $28\times 28$ reconstructions}
\end{algorithmic}
\end{algorithm}

\subsection*{Fashion-MNIST Super-resolution: More Examples}
\label{subsec:more_examples}

\begin{figure*}[ht]
\centering
\includegraphics[width=1.0\textwidth]{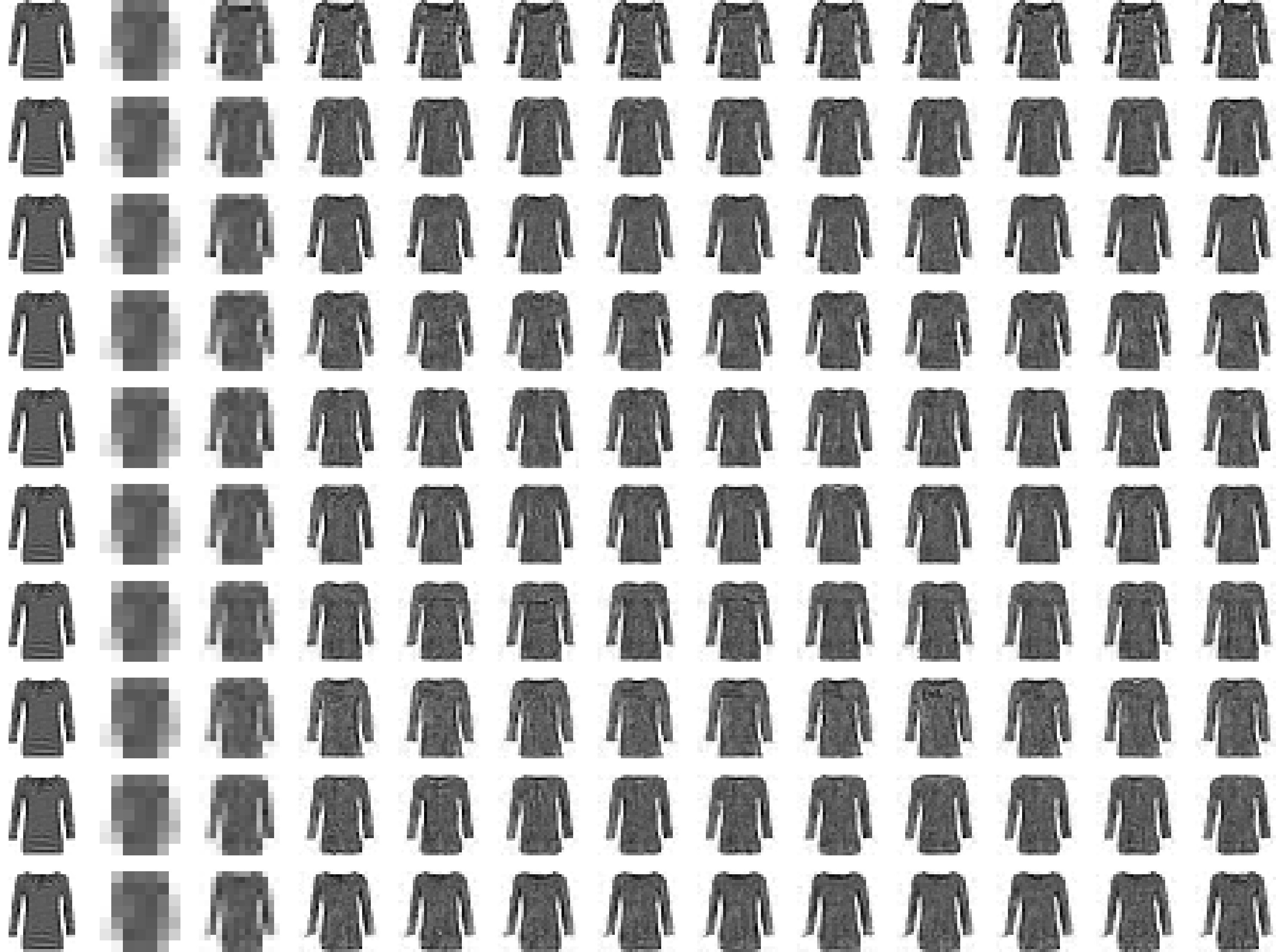}
\caption{Two-scale super-resolution of an example from the fashion-MNIST T-shirt class. Column 1: original image, $\mathbf{x}'_0$. Column 2: $4\times 4$-downsampled input to the model, $\mathbf{y}$. Column 3: 10 different reconstructions of the image at $2\times 2$-downsampled resolution, $\mathbf{x}_{1,i}'$. Columns 4-13: 10 different reconstructions of the image at its original resolution for each of the intermediate reconstructions in column 3, $\mathbf{x}_{0,ij}'$.}
\label{fig:T_shirts}
\end{figure*}

\begin{figure*}[h]
\centering
\includegraphics[width=1.0\textwidth]{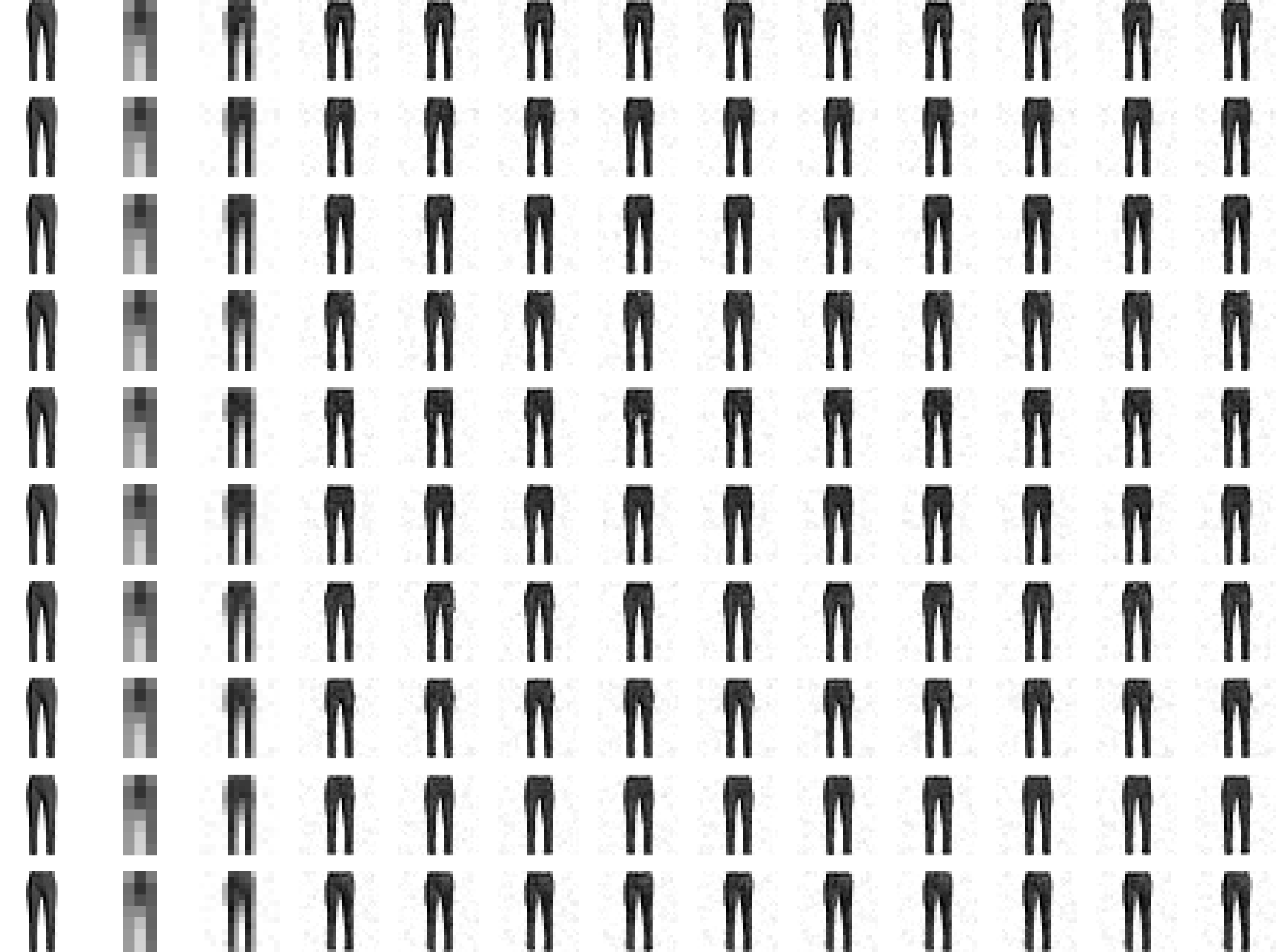}
\caption{Two-scale super-resolution of an example from the fashion-MNIST trouser class. Column 1: original image, $\mathbf{x}'_0$. Column 2: $4\times 4$-downsampled input to the model, $\mathbf{y}$. Column 3: 10 different reconstructions of the image at $2\times 2$-downsampled resolution, $\mathbf{x}_{1,i}'$. Columns 4-13: 10 different reconstructions of the image at its original resolution for each of the intermediate reconstructions in column 3, $\mathbf{x}_{0,ij}'$.}
\label{fig:trousers}
\end{figure*}

\begin{figure*}[h]
\centering
\includegraphics[width=1.0\textwidth]{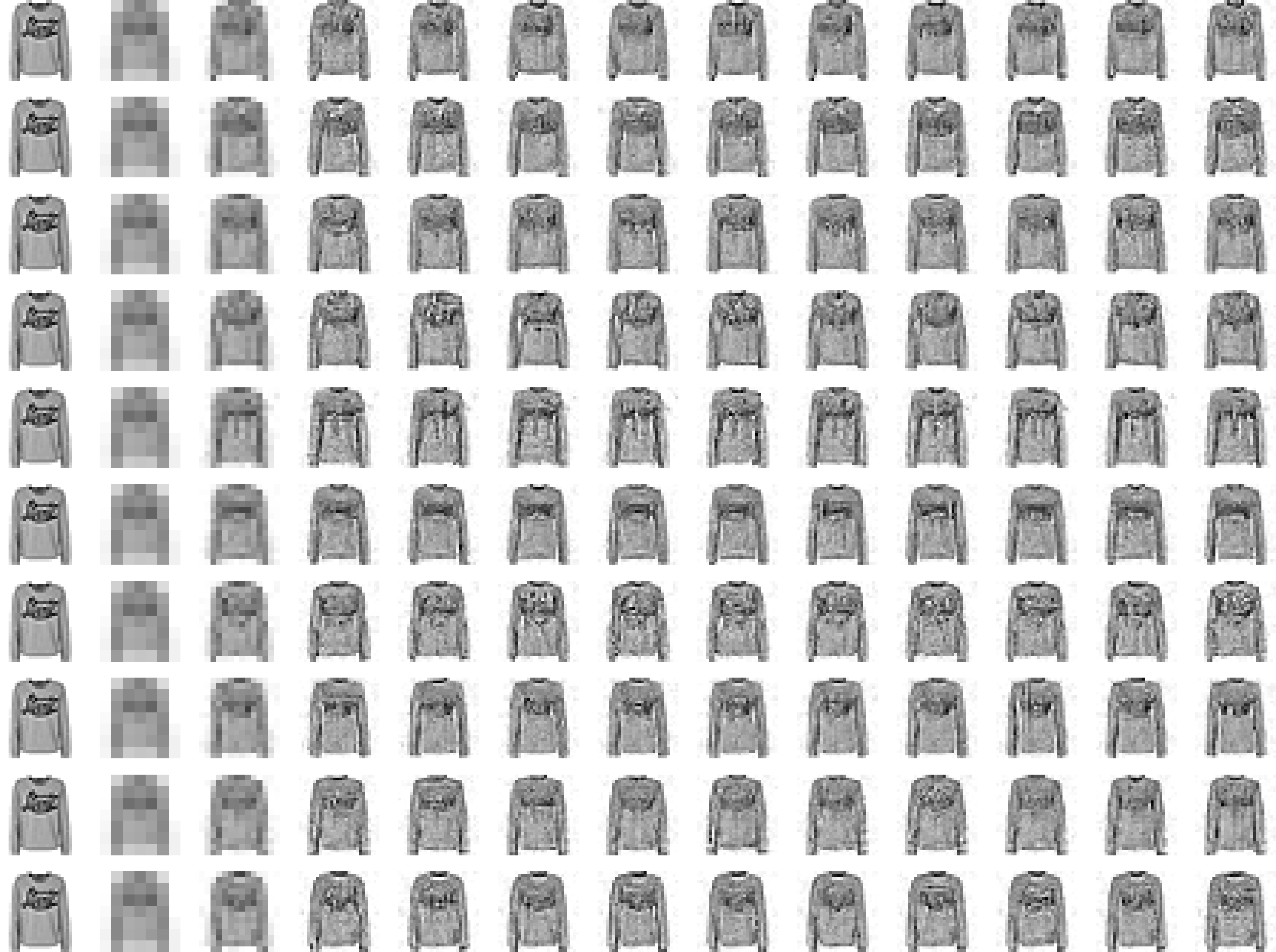}
\caption{Two-scale super-resolution of an example from the fashion-MNIST pullover class. Column 1: original image, $\mathbf{x}'_0$. Column 2: $4\times 4$-downsampled input to the model, $\mathbf{y}$. Column 3: 10 different reconstructions of the image at $2\times 2$-downsampled resolution, $\mathbf{x}_{1,i}'$. Columns 4-13: 10 different reconstructions of the image at its original resolution for each of the intermediate reconstructions in column 3, $\mathbf{x}_{0,ij}'$.}
\label{fig:pullovers}
\end{figure*}

\begin{figure*}[h]
\centering
\includegraphics[width=1.0\textwidth]{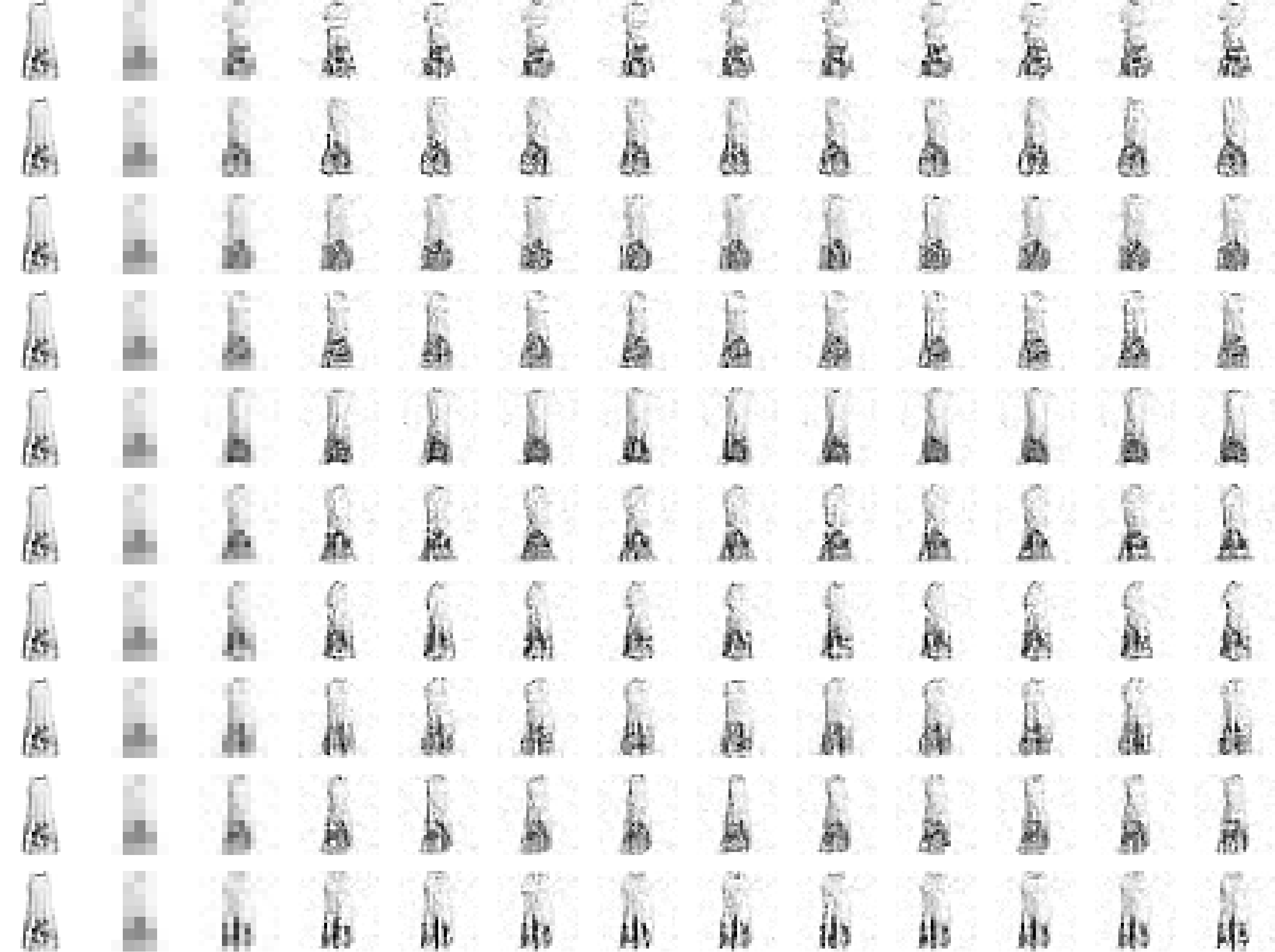}
\caption{Two-scale super-resolution of an example from the fashion-MNIST dress class. Column 1: original image, $\mathbf{x}'_0$. Column 2: $4\times 4$-downsampled input to the model, $\mathbf{y}$. Column 3: 10 different reconstructions of the image at $2\times 2$-downsampled resolution, $\mathbf{x}_{1,i}'$. Columns 4-13: 10 different reconstructions of the image at its original resolution for each of the intermediate reconstructions in column 3, $\mathbf{x}_{0,ij}'$.}
\label{fig:dresses}
\end{figure*}

\begin{figure*}[h]
\centering
\includegraphics[width=1.0\textwidth]{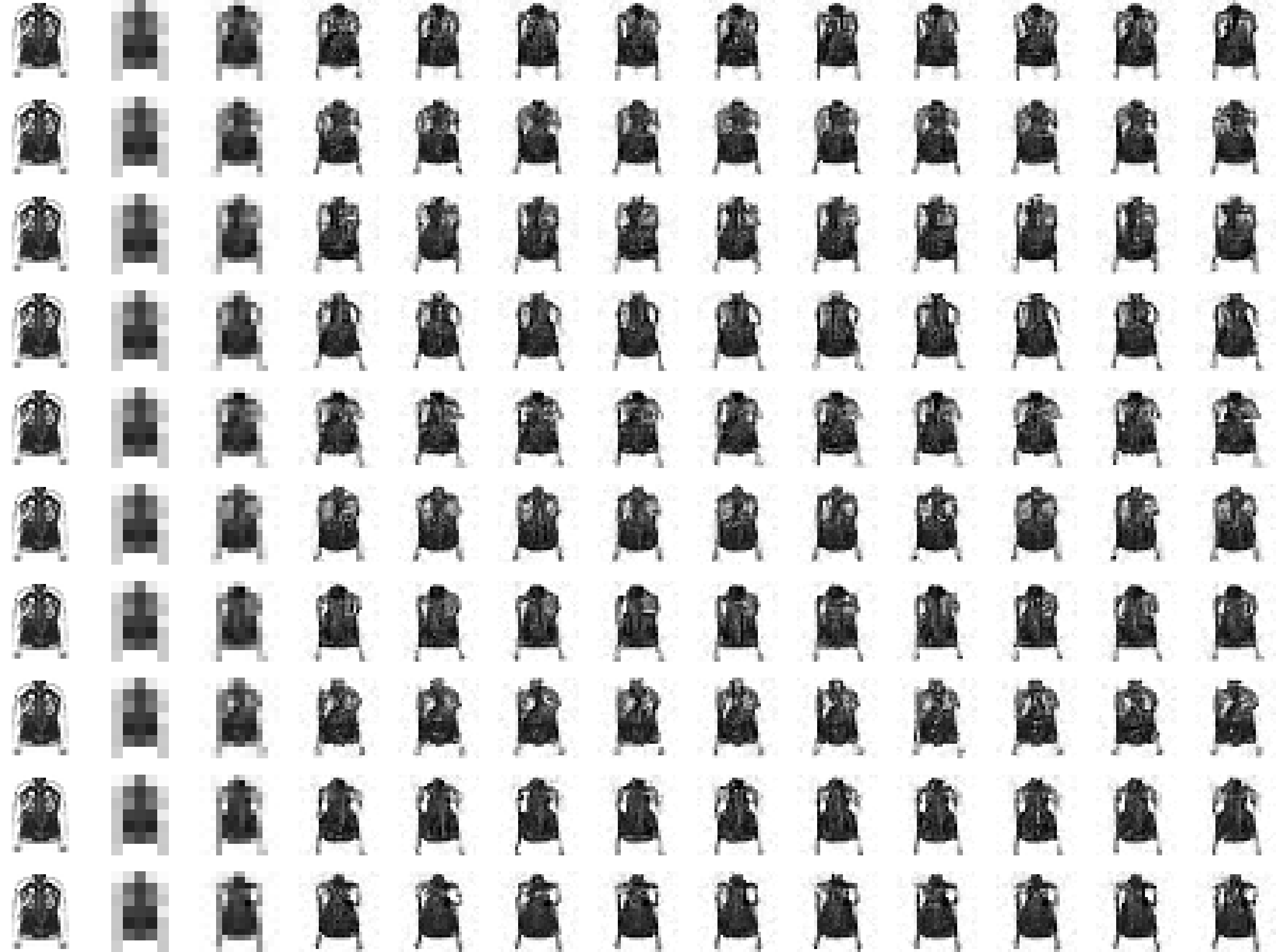}
\caption{Two-scale super-resolution of an example from the fashion-MNIST coat class. Column 1: original image, $\mathbf{x}'_0$. Column 2: $4\times 4$-downsampled input to the model, $\mathbf{y}$. Column 3: 10 different reconstructions of the image at $2\times 2$-downsampled resolution, $\mathbf{x}_{1,i}'$. Columns 4-13: 10 different reconstructions of the image at its original resolution for each of the intermediate reconstructions in column 3, $\mathbf{x}_{0,ij}'$.}
\label{fig:coats}
\end{figure*}

\begin{figure*}[h]
\centering
\includegraphics[width=1.0\textwidth]{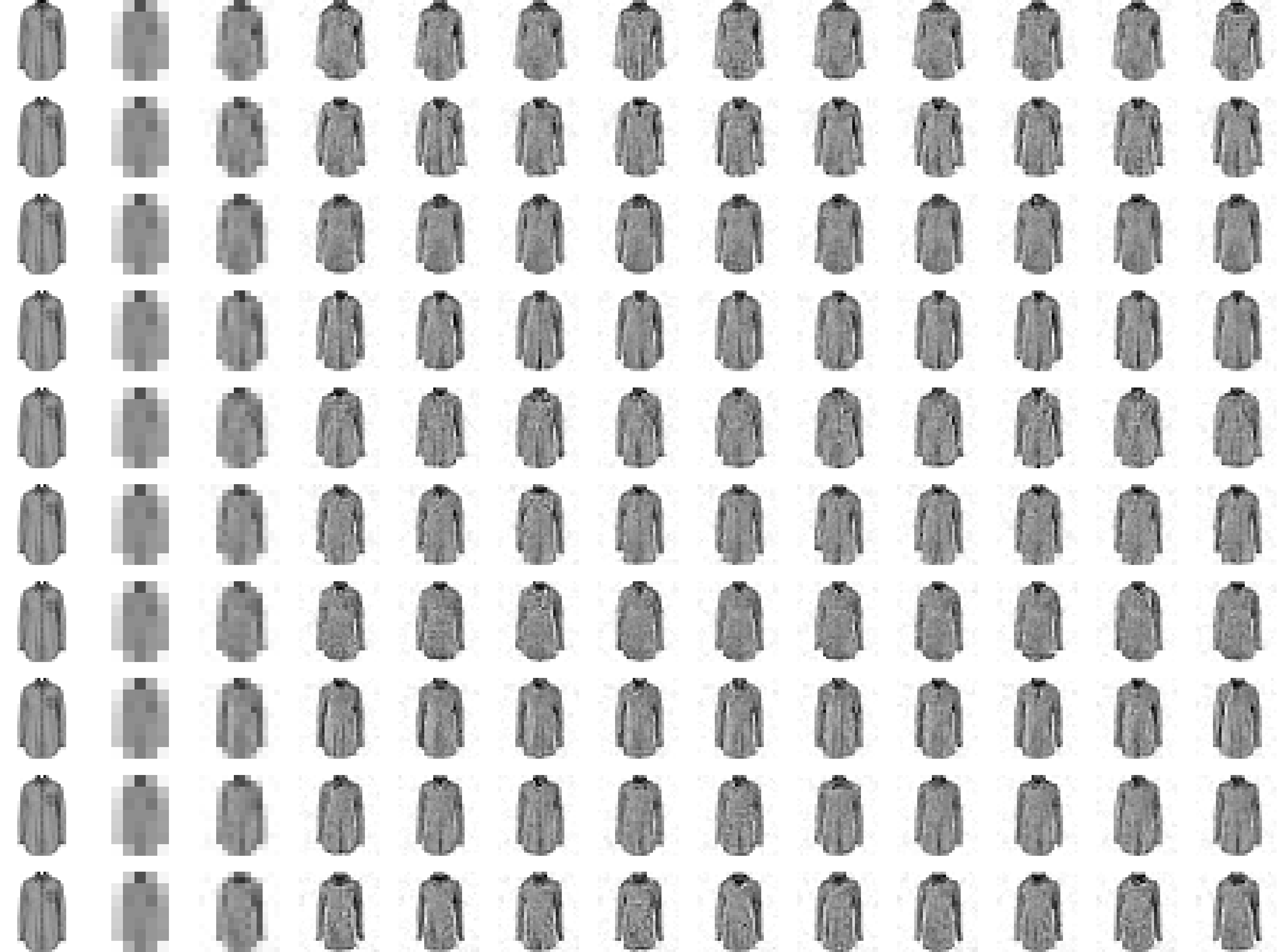}
\caption{Two-scale super-resolution of an example from the fashion-MNIST shirt class. Column 1: original image, $\mathbf{x}'_0$. Column 2: $4\times 4$-downsampled input to the model, $\mathbf{y}$. Column 3: 10 different reconstructions of the image at $2\times 2$-downsampled resolution, $\mathbf{x}_{1,i}'$. Columns 4-13: 10 different reconstructions of the image at its original resolution for each of the intermediate reconstructions in column 3, $\mathbf{x}_{0,ij}'$.}
\label{fig:shirts}
\end{figure*}

\begin{figure*}[h]
\centering
\includegraphics[width=1.0\textwidth]{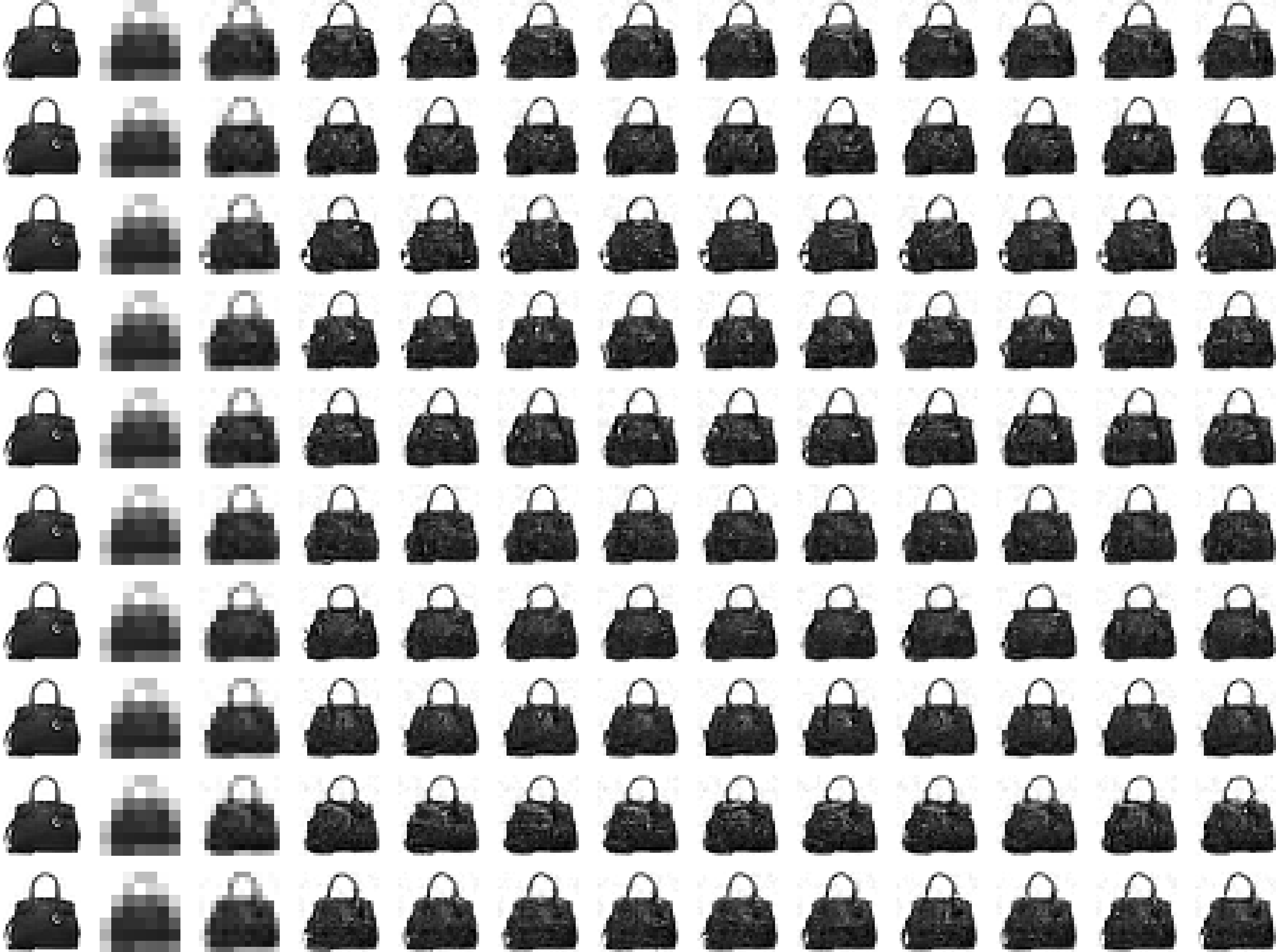}
\caption{Two-scale super-resolution of an example from the fashion-MNIST bag class. Column 1: original image, $\mathbf{x}'_0$. Column 2: $4\times 4$-downsampled input to the model, $\mathbf{y}$. Column 3: 10 different reconstructions of the image at $2\times 2$-downsampled resolution, $\mathbf{x}_{1,i}'$. Columns 4-13: 10 different reconstructions of the image at its original resolution for each of the intermediate reconstructions in column 3, $\mathbf{x}_{0,ij}'$.}
\label{fig:bags}
\end{figure*}

\begin{figure*}[h]
\centering
\includegraphics[width=1.0\textwidth]{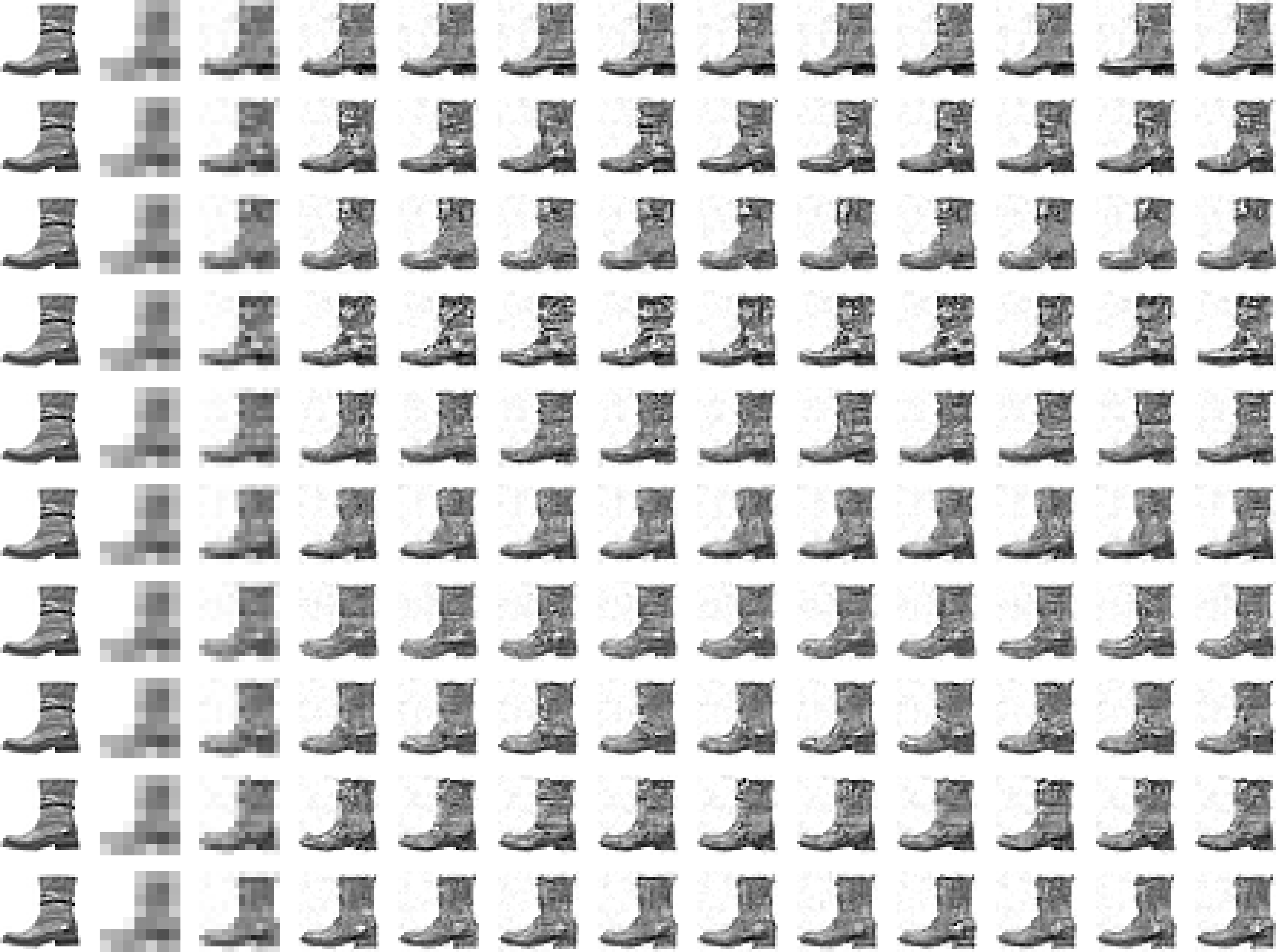}
\caption{Two-scale super-resolution of an example from the fashion-MNIST ankle boot class. Column 1: original image, $\mathbf{x}'_0$. Column 2: $4\times 4$-downsampled input to the model, $\mathbf{y}$. Column 3: 10 different reconstructions of the image at $2\times 2$-downsampled resolution, $\mathbf{x}_{1,i}'$. Columns 4-13: 10 different reconstructions of the image at its original resolution for each of the intermediate reconstructions in column 3, $\mathbf{x}_{0,ij}'$.}
\label{fig:ankle_boots}
\end{figure*}

The models $F_1\circ F_2$ described in Section 6.4 are trained on the full fashion-MNIST dataset, including all 10 classes, \textit{without} class labels. The paper includes examples from the sneakers and sandals classes; Figs.\ \ref{fig:T_shirts}-\ref{fig:ankle_boots} show one example each from the other 8 classes, following the same format. Generally, the model is fairly good at sampling textures and large-scale features such as buckles, collars, or seams. It performs poorly on rarer semantic content such as text. Also note that several classes (particularly the trousers class) are extremely homogeneous over the dataset; most of the variation in reconstructions is packed into the 2\% noise added to the training data. Much more so than the continuous toy dataset, there is no reason to think the choice of coupling law (affine scaling transformation) is well-suited to the structure of this dataset. The learned distributions thus likely approximate the true distribution, but the model has probably not learned a good representation of the underlying structure of the problem. In practical terms, this means rare semantics (such as those found in Fig.\ \ref{fig:pullovers}) are not likely to be sampled during reconstruction.

\bibliography{bibliography}